\documentclass[10pt,twocolumn,letterpaper]{article}
\pdfoutput=1

\usepackage{cvpr}
\usepackage{times}
\usepackage{epsfig}
\usepackage{graphicx}
\usepackage{amsmath}
\usepackage{amssymb}
\usepackage{flushend}
\usepackage{subcaption}
\usepackage{multirow}
\usepackage{bm}

\usepackage[bottom,flushmargin,hang,multiple]{footmisc}

\usepackage{amsmath,amsthm}
\usepackage{amssymb, color}
\usepackage{float}
\newtheorem{remark}{Remark}[section]

\graphicspath{{figure/}}

\renewcommand{\phi}{\varphi}

\renewcommand{\hat}{\widehat}

\newcommand{\R}{\mathbb{R}}

\usepackage{mathtools}

\usepackage[pagebackref=false,breaklinks=true,letterpaper=true,colorlinks,bookmarks=false]{hyperref}

\newcommand\blfootnote[1]{%
  \begingroup
  \renewcommand\thefootnote{}\footnote{#1}%
  \addtocounter{footnote}{-1}%
  \endgroup
}

\cvprfinalcopy % *** Uncomment this line for the final submission

\def\cvprPaperID{6172} % *** Enter the CVPR Paper ID here

\ifcvprfinal\fi
\begin{document}

%%%%%%%%% TITLE
\title{Object-driven Text-to-Image Synthesis via Adversarial Training}

\author{
  $\textrm{Wenbo Li}^{\dag*1,2} \quad \textrm{Pengchuan Zhang}^{*2} \quad \textrm{Lei Zhang}^{3}$ \\ $\textrm{Qiuyuan Huang}^{2} \quad \textrm{Xiaodong He}^{4} \quad \textrm{Siwei Lyu}^{1} \quad \textrm{Jianfeng Gao}^{2}$\\
  %\and
  $\textrm{}^{1}$University at Albany, SUNY$\quad\textrm{}^{2}$Microsoft Research AI$\quad\textrm{}^{3}$Microsoft$\quad\textrm{}^{4}$JD AI Research\\
  {\tt\small $\{$wli20,slyu$\}$@albany.edu}, {\tt\small $\{$penzhan,leizhang,qihua,jfgao$\}$@microsoft.com}, {\tt\small xiaodong.he@jd.com}
}

\maketitle
\blfootnote{$\dag$ Work was performed when was an intern with Microsoft Research AI.}
\blfootnote{* indicates equal contributions.}
\begin{abstract}
  In this paper, we propose Object-driven Attentive Generative Adversarial Newtorks (Obj-GANs) that allow object-centered text-to-image synthesis for complex scenes. Following the two-step (layout-image) generation process, a novel object-driven attentive image generator is proposed to synthesize salient objects by paying attention to the most relevant words in the text description and the pre-generated semantic layout. In addition, a new Fast R-CNN based object-wise discriminator is proposed to provide rich object-wise discrimination signals on whether the synthesized object matches the text description and the pre-generated layout. The proposed Obj-GAN significantly outperforms the previous state of the art in various metrics on the large-scale COCO benchmark, increasing the Inception score by 27\% and decreasing the FID score by 11\%. A thorough comparison between the traditional grid attention and the new object-driven attention is provided through analyzing their mechanisms and visualizing their attention layers, showing insights of how the proposed model generates complex scenes in high quality.
\end{abstract}

\vspace{-5mm}
\section{Introduction}
\label{sec:intro}
Synthesizing images from text descriptions (known as {\em Text-to-Image synthesis}) is an important machine learning task, which requires handling ambiguous and incomplete information in natural language descriptions and learning across vision and language modalities.
Approaches based on Generative Adversarial Networks (GANs)~\cite{goodfellow2014generative} have recently achieved promising results on this task~\cite{reed2016generative,reed2016learning,Han16stackgan,Han17stackgan2,xu2017attngan,ma2018gan,hong2018inferring,johnson2018image,zhang2018text}.
%In the GAN-based methods, a discriminator network is introduced to play an adversarial game with the text-to-image generator: the discriminator tries to differentiate the generated text-image pairs from the real data, while the generator tries to fool the discriminator by generating more realistic images matching the given text description.
Most GAN based methods synthesize the image conditioned only on a global sentence vector, which may miss important fine-grained information at the word level, and prevents the generation of high-quality images. More recently, AttnGAN~\cite{xu2017attngan} is proposed which introduces the attention mechanism \cite{XuBKCCSZB15,YangHGDS16,Dzmitry14,Ashish17} into the GAN framework, thus allows attention-driven, multi-stage refinement for fine-grained text-to-image generation.

\begin{figure}[tb]
\begin{center}
\includegraphics[width=0.95\linewidth]{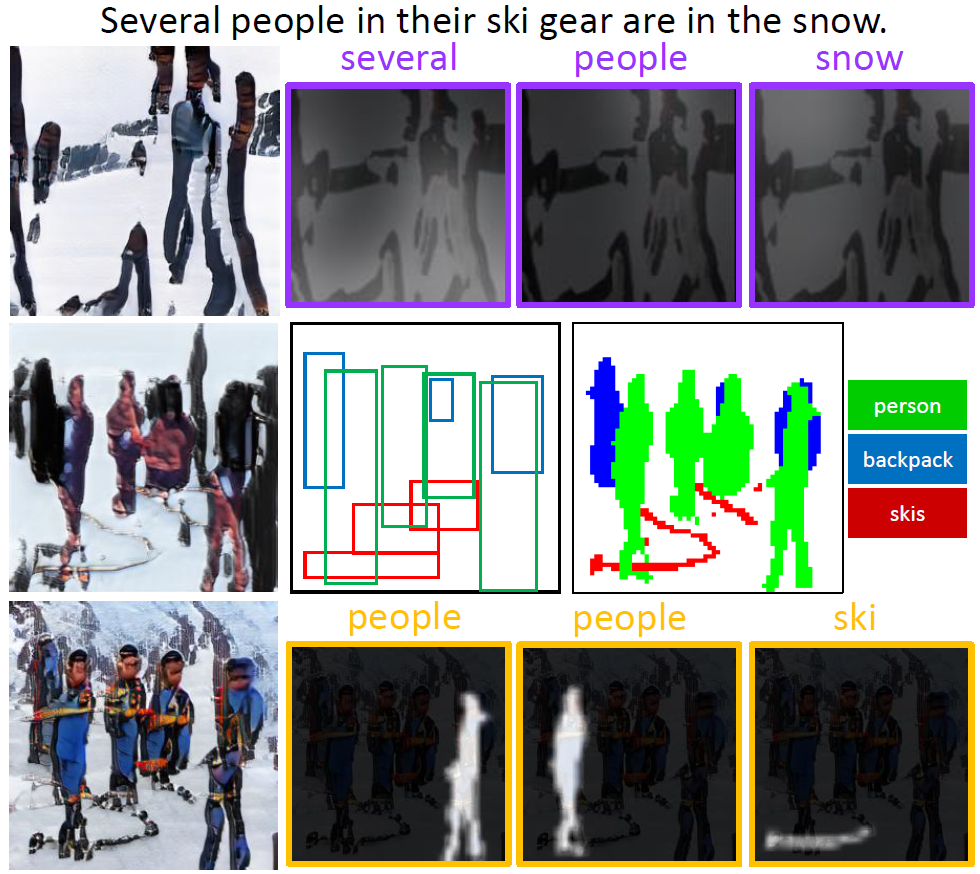}
\end{center}
\vspace{-15pt}
   \caption{\small Top: AttnGAN~\cite{xu2017attngan} and its grid attention visualization. Middle: our modified implementation of two-step (layout-image) generation proposed in~\cite{hong2018inferring}. Bottom: our Obj-GAN and its object-driven attention visualization. The middle and bottom generations use the same {\it generated} semantic layout, and the only difference is the object-driven attention.}
\vspace{-16pt}
	\label{fig:maineg}
\end{figure}

Although images with realistic texture have been synthesized on simple datasets, such as birds~\cite{xu2017attngan,ma2018gan} and flowers~\cite{Han17stackgan2}, most existing approaches do not specifically model objects and their relations in images and thus have difficulties in generating complex scenes such as those in the COCO dataset~\cite{LinMBHPRDZ14}. For example, generating images from a sentence ``several people in their ski gear are in the snow'' requires modeling of different objects (people, ski gear) and their interactions (people on top of ski gear), as well as filling the missing information (\eg, the rocks in the background). In the top row of Fig.~\ref{fig:maineg}, the image generated by AttnGAN does contain scattered texture of people and snow, but the shape of people are distorted and the picture's layout is semantically not meaningful. \cite{hong2018inferring} remedies this problem by first constructing a semantic layout from the text and then synthesizing the image by a deconvolutional image generator. However, the fine-grained word/object-level information is still not explicitly used for generation. Thus, the synthesized images do not contain enough details to make them look realistic (see the middle row of Fig.~\ref{fig:maineg}).

In this study, we aim to generate high-quality complex images with semantically meaningful layout and realistic objects.
To this end, we propose a novel Object-driven Attentive Generative Adversarial Networks (Obj-GAN) that effectively capture and utilize fine-grained word/object-level information for text-to-image synthesis. %to generate high-quality complex scenes.
The Obj-GAN consists of a pair of object-driven attentive image generator and object-wise discriminator, and a new object-driven attention mechanism. The proposed image generator takes as input the text description and a pre-generated semantic layout and synthesize high-resolution images via multiple-stage coarse-to-fine process. At {\it every} stage, the generator synthesizes the image region within a bounding box by focusing on words that are most relevant to the object in that bounding box, as illustrated in the bottom row of Fig.~\ref{fig:maineg}. More specifically, using a new object-driven attention layer, it uses the class label to query words in the sentences to form a word context vector, as illustrated in Fig.~\ref{fig:attention}, and then synthesizes the image region conditioned on the class label and word context vector. The object-wise discriminator checks every bounding box to make sure that the generated object indeed matches the pre-generated semantic layout. To compute the discrimination losses for all bounding boxes simultaneously and efficiently, our object-wise discriminator is based on a Fast R-CNN~\cite{girshick2015fast}, with a binary cross-entropy loss for each bounding box.

The contribution of this work is three-folded.
(\textit{i}) An Object-driven Attentive  Generative Network (Obj-GAN) is proposed for synthesizing complex images from text descriptions. Specifically, two novel components are proposed, including the object-driven attentive generative network and the object-wise discriminator. %Obj-GAN is able to  composed of a multi-stage object-level attentive generative network and an object-level discriminator.
(\textit{ii}) Comprehensive evaluation on a large-scale COCO benchmark shows that our Obj-GAN significantly outperforms previous state-of-the-art text-to-image synthesis methods. Detailed ablation study is performed to empirically evaluate the effect of different components in Obj-GAN.
(\textit{iii}) A thorough analysis is performed through visualizing the attention layers of the Obj-GAN, showing insights of how the proposed model generates complex scenes in high quality. Compared with the previous work, our object-driven attention is more robust and interpretable, and significantly improves the object generation quality in complex scenes.

\begin{figure*}[tb]
\begin{center}
\includegraphics[width=1\linewidth]{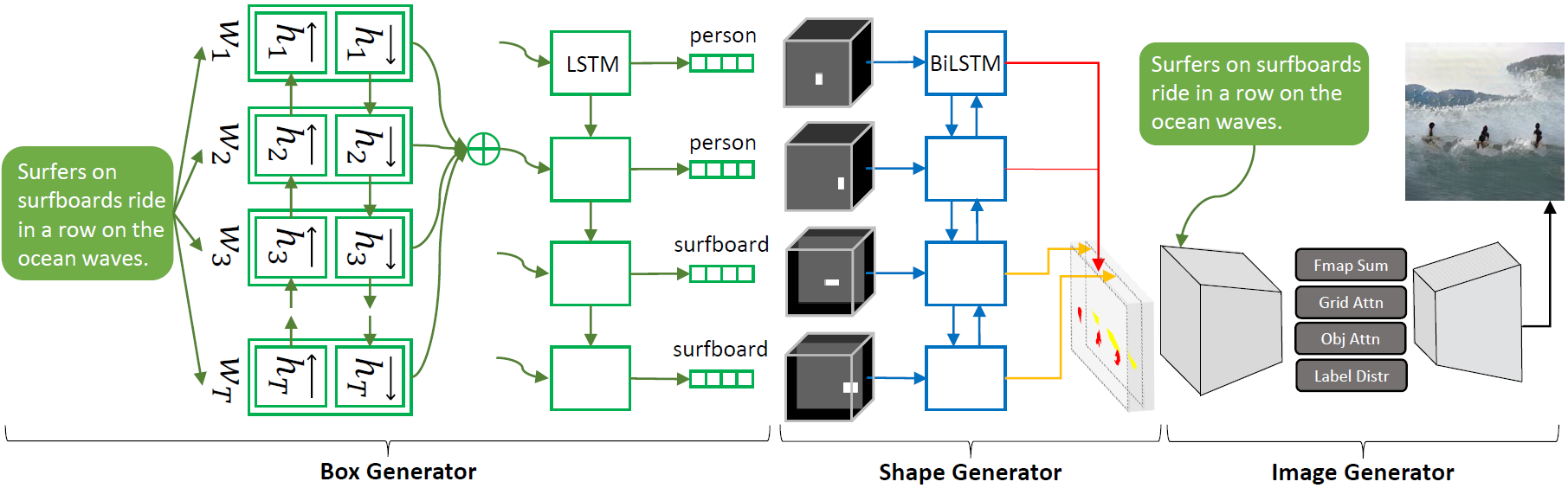}
\end{center}
\vspace{-16pt}
   \caption{\small Obj-GAN completes the text-to-image synthesis in two steps: the layout generation and the image generation. The layout generation contains a bounding box generator and a shape generator. The image generation uses the object-driven attentive image generator.}
\vspace{-18pt}
	\label{fig:3steps}
\end{figure*}

\section{Related Work}
\label{sec:related}
Generating photo-realistic images from text descriptions, though challenging, is important to many real-world applications such as art generation and computer-aided design. There has been much research effort for this task through different approaches, such as variational inference~\cite{MansimovPBS15, Gregor15DRAW}, approximate Langevin process~\cite{Reed17parallel}, conditional PixelCNN via maximal likelihood estimation~\cite{Oord16, Reed17parallel}, and conditional generative adversarial networks~\cite{reed2016generative, reed2016learning, Han16stackgan, Han17stackgan2}. Compared with other approaches, Generative Adversarial Networks (GANs)~\cite{goodfellow2014generative} have shown better performance in image generation~\cite{Radford15, DentonCSF15, Salimans2016, Christian2016, pix2pix2017,huang2018turbo}. However, existing GAN based text-to-image synthesis is usually conditioned only on the global sentence vector, which misses important fine-grained information at the word level, and thus lacks the ability to generate high-quality images. \cite{xu2017attngan} uses the traditional grid visual attention mechanism in this task, which enables synthesizing fine-grained details at different image regions by paying attentions to the relevant words in the text description.

To explicitly encode the semantic layout into the generator, \cite{hong2018inferring} proposes to decompose the generation process into two steps, in which it first constructs a semantic layout (bounding boxes and object shapes) from the text and then synthesizes an image conditioned on the layout and text description. \cite{johnson2018image} also proposes such a two-step process to generate images from scene graphs, and their process can be trained end-to-end. In this work, the proposed Obj-GAN follows the two-step generation process as \cite{hong2018inferring}. However, \cite{hong2018inferring} encodes the text into a single global sentence vector, which loses word-level fine-grained information. Moreover, it uses the image-level GAN loss for the discriminator, which is less effective at providing object-wise discrimination signal for generating salient objects. We propose a new object-driven attention mechanism to provide fine-grained information (words in the text description and objects in the layout) for different components, including an attentive seq2seq bounding box generator, an attentive image generator and an object-wise discriminator.

The attention mechanism has recently become a crucial part of vision-language multi-modal intelligence tasks. The traditional grid attention mechanism has been successfully used in modeling multi-level dependencies in image captioning~\cite{XuBKCCSZB15}, image question answering~\cite{YangHGDS16}, text-to-image generation~\cite{xu2017attngan}, unconditional image synthesis~\cite{zhang2018self} and image-to-image translation~\cite{ma2018gan}, image/text retrieval~\cite{Lee2018Stacked}. In 2018, \cite{anderson2017bottom} proposes a bottom-up attention mechanism, which enables attention to be calculated over semantic meaningful regions/objects in the image, for image captioning and visual question-answering. Inspired by these works, we propose Obj-GAN which for the first time develops an object-driven attentive generator plus an object-wise discriminator, thus enables GANs to synthesize high-quality images of complicated scenes.

\section{Object-driven Attentive GAN}
\label{sec:model}
As illustrated in Fig.~\ref{fig:3steps}, the Obj-GAN performs text-to-image synthesis in two steps: generating a semantic layout (class labels, bounding boxes, shapes of salient objects), and then generating the image. In the image generation step, the object-driven attentive generator and object-wise discriminator are designed to enable image generation conditioned on the semantic layout generated in the first step.

The input of Obj-GAN is a sentence with $T_s$ tokens. With a pre-trained bi-LSTM model, we encode its words as word vectors $e \in \R^{D\times T_s}$ and the entire sentence as a global sentence vector $\bar{e}\in \R^{D}$. We provide details of this pre-trained bi-LSTM model and the implementation details of other modules of Obj-GAN in $\S$~\ref{sec:appendix}.

% In section~\ref{subsec:bbox}, we present our attentive bounding box generator and shape generator. In section~\ref{subsec:imagegenerator}, we present our attentive multistage image generator, which first generates a low-resolution image conditioned on the global sentence vector and shapes of salient objects and then refines details in different regions by paying attention to most relevant words and pre-generated class labels. In section~\ref{subsec:imageloss}, we present discriminators that provide both patch-wise GAN loss and object-wise GAN loss. In section~\ref{subsec:metrics}, we present metrics to evaluate qualities of synthesized images.
% we use a deep attentional multi-modal similarity model (DAMSM) to train a joint embedding for words and image regions, which is used in the rest of the model. The DAMSM loss also provides an additional fine-grained image-text matching loss for training the image generator.

\subsection{Semantic layout generation}
\label{subsec:bbox}
In the first step, the Obj-GAN takes the sentence as input and generates a semantic layout, a sequence of objects specified by their bounding boxes (with class labels) and shapes. As illustrated in Fig.~\ref{fig:3steps}, a box generator first generates a sequence of bounding boxes, and then a shape generator generates their shapes. This part resembles the bounding box generator and shape generator in \cite{hong2018inferring}, and we put our implementation details in $\S$~\ref{sec:appendix}.

\noindent{\bf Box generator.} We train an attentive seq2seq model~\cite{Dzmitry14}, also referring to Fig.~\ref{fig:3steps}, as the box generator:
\begin{equation}\label{eqn:boxgen}
    B_{1:T} := [B_1, B_2, \dots, B_T] \sim G_{\text{box}}(e).
\end{equation}
Here, $e$ are the pre-trained bi-LSTM word vectors, $B_t = (l_t, b_t)$ are the class label of the $t$'s object and its bounding box $b = (x, y, w, h) \in \R^4$. In the rest of the paper, we will also call the label-box pair $B_t$ as a bounding box when no confusion arises. Since most of the bounding boxes have corresponding words in the sentence, the attentive seq2seq model captures this correspondence better than the seq2seq model used in \cite{hong2018inferring}.
%In the test time, we randomly sample boxes from the model and recursively feed the current box back into the model to sample the next one. We terminate the generation when an ``End-Of-Sequence'' token is emitted or the length of the sequence reaches 10.

\noindent{\bf Shape generator.} Given the bounding boxes $B_{1:T}$, the shape generator predicts the shape of each object in its bounding box, \ie,
\begin{equation}\label{eqn:hmapgen}
    \hat{M}_{1:T} = G_{\text{shape}}(B_{1:T}, z_{1:T}).
\end{equation}
where $z_{t} \sim \mathcal{N}(0,1)$ is a random noise vector. Since the generated shapes not only need to match the location and category information provided by $B_{1:T}$, but also should be aligned with its surrounding context, we build $G_{\text{shape}}$ based on a bi-directional convolutional LSTM, as illustrated in Fig.~\ref{fig:3steps}.
%Similar to most generative networks, the architecture of $G_{\text{shape}}$ follows the encoder-decoder fashion, and models the contextual information by a Bi-convLSTM in the latent space.
Training of $G_{\text{shape}}$ is based on the GAN framework \cite{hong2018inferring}, in which a perceptual loss is also used to constrain the generated shapes and to stabilize the training.

\begin{figure*}[tb]
\begin{center}
\includegraphics[width=1\linewidth]{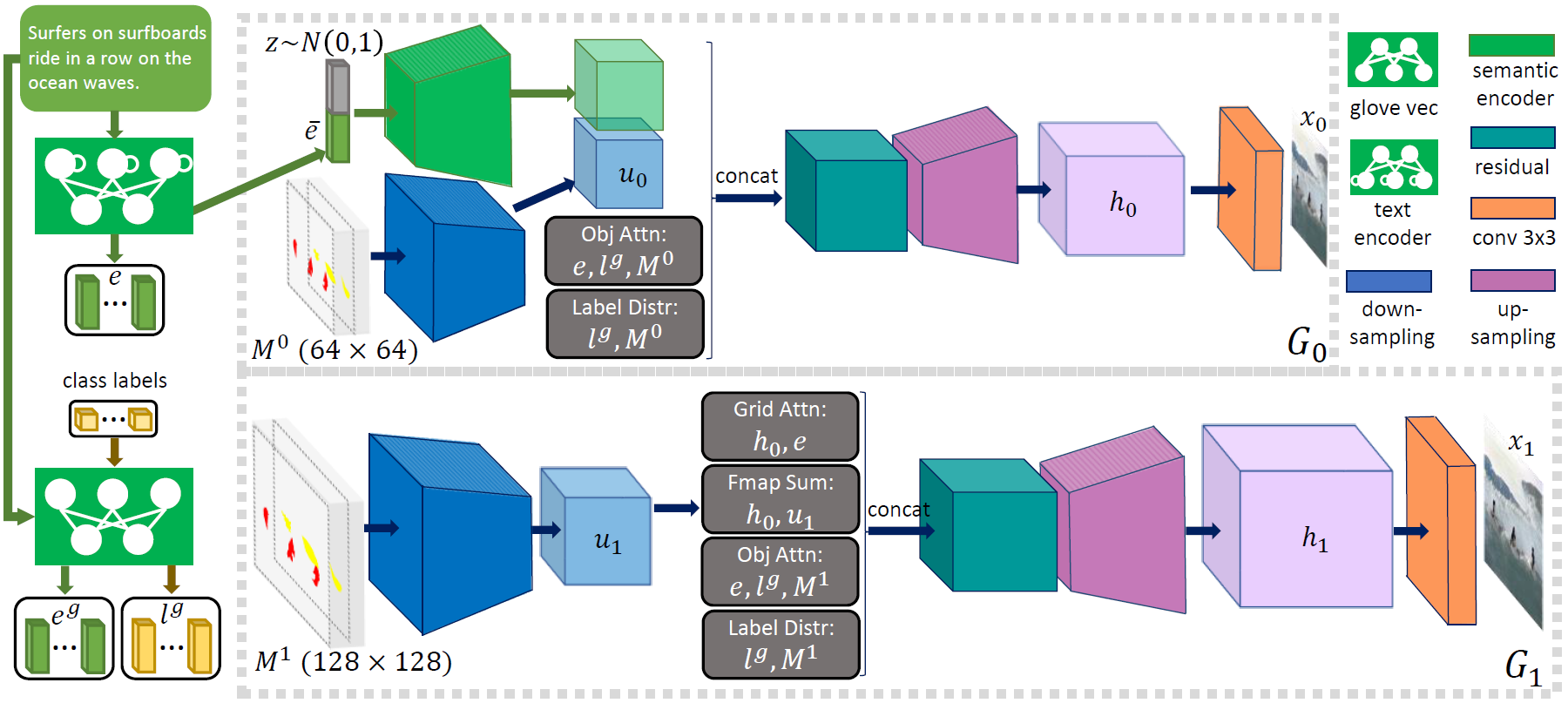}
\end{center}
\vspace{-16pt}
   \caption{\small The object-driven attentive image generator.}
\vspace{-19pt}
	\label{fig:ImageGenerator}
\end{figure*}

\subsection{Image generation}
\subsubsection{Attentive multistage image generator}
\label{subsec:imagegenerator}
As shown in Fig.~\ref{fig:ImageGenerator}, the proposed attentive multistage generative network has two generators ($G_0, G_1$). The base generator $G_0$ first generates a low-resolution image $\hat{x}_0$ conditioned on the global sentence vector and the pre-generated semantic layout. The refiner $G_1$ then refines details in different regions by paying attention to most relevant words and pre-generated class labels and generates a higher resolution image $\hat{x}_1$. Specifically,
  \vspace{-2mm}
  \begin{equation*}
   \begin{aligned}
    h_0 &= F_0(z, \quad \Bar{e}, \quad\text{Enc}(M^0), c^{\text{obj}}, c^{\text{lab}}), \quad \hat{x}_0 = G_0(h_0),\\
    h_1 &= F_1(c^{\text{pat}}, h_{0}+\text{Enc}(M^1), c^{\text{obj}}, c^{\text{lab}}), \quad \hat{x}_1 = G_1(h_1),
   \end{aligned}
   \vspace{-1mm}
  \end{equation*}
where (\textit{i}) $z$ is a random vector with standard normal distribution; (\textit{ii}) $\text{Enc}(M^0)$ ( $\text{Enc}(M^1)$ ) is the encoding of low-resolution shapes $M^0$ (higher-resolution shapes $M^1$); (\textit{iii}) $c^{\text{pat}}= F_{\text{attn}}^{\text{grid}}(e,h_0)$ are the patch-wise context vectors from the traditional grid attention, (\textit{iv}) $c^{\text{obj}}=F_{\text{attn}}^{\text{obj}}(e, e^g, l^g, M)$ are the object-wise context vectors from our new object-driven attention, and $c^{\text{lab}}=c^{\text{lab}}(l^g, M)$ are the label context vectors from class labels. We can stack more refiners to the generation process and get higher and higher resolution images. In this paper, we have two refiners ($G_1$ and $G_2$) and finally generate images with resolution $256\times 256$.
 %See Fig.~\ref{fig:attention} for illustration of these two kinds of image regions.
\begin{figure}[tb]
\begin{center}
\includegraphics[width=1\linewidth]{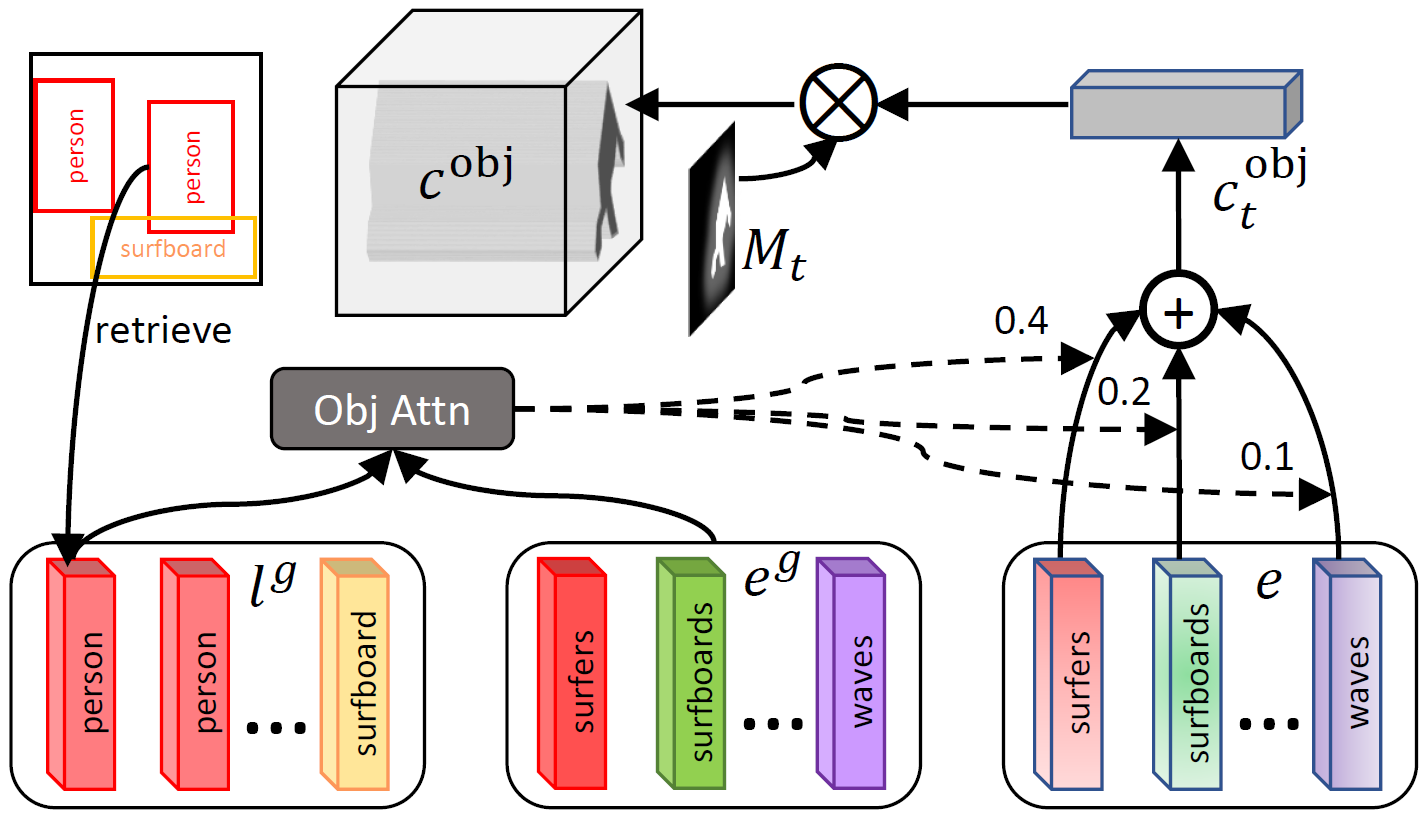}
\end{center}
\vspace{-16pt}
   \caption{\small Object-driven attention.}
\vspace{-18pt}
	\label{fig:attention}
\end{figure}

\noindent{\bf Compute context vectors via attention.} Both patch-wise context vectors $c^{\text{pat}}$ and object-wise context vectors $c^{\text{obj}}$ are attention-driven context vectors for specific image regions, and encode information from the words that are most relevant to that image region. Patch-wise context vectors are for uniform-partitioned image patches determined by the uniform down-sampling/up-sampling structure of CNN, but these patches are not semantically meaningful. Object-wise context vectors are for semantically meaningful image regions specified by bounding boxes, but these regions are at different scales and may have overlaps.

Specifically, the patch-wise context vector $c_j^{\text{pat}}$ ( objective-wise context vector $c_t^{\text{obj}}$) is a dynamic representation of word vectors relevant to patch $j$ (bounding box $B_t$), which is calculated by
\vspace{-3mm}
 \begin{equation}\label{eq:AttnGAN}
  c_j^{\text{pat}} = \sum_{i=1}^{T_s}\beta_{j,i}^{\text{pat}} e_{i}, \quad
  c_t^{\text{obj}} = \sum_{i=1}^{T_s}\beta_{t,i}^{\text{obj}} e_{i}.
 \vspace{-3mm}
 \end{equation}
Here, $\beta_{j,i}^{\text{pat}}$ ( $\beta_{t,i}^{\text{obj}}$ ) indicates the weight the model attends to the $i$'th word when generating patch $j$ (bounding box $B_t$) and is computed by
 \vspace{-3mm}
 \begin{align}
  \beta_{j,i}^{\text{pat}} = \frac{\exp(s_{j,i}^{\text{pat}})}{\sum_{k=1}^{T_s}{\exp(s_{j,k}^{\text{pat}})}},  \; \; \quad \; s_{j,i}^{\text{pat}} = (h_j)^T e_{i}, \label{eq:AttnGANbeta_pat}\\
  \beta_{t,i}^{\text{obj}} = \frac{\exp(s_{t,i}^{\text{obj}})}{\sum_{k=1}^{T_s}{\exp(s_{t,k}^{\text{obj}})}},  \; \; \quad \; s_{t,i}^{\text{obj}} = (l_t^g)^T e^g_{i}. \label{eq:AttnGANbeta_inst}
   \vspace{-3mm}
 \end{align}
For the traditional grid attention, we use the image region feature $h_j$, which is one column in the previous hidden layer $h \in \R^{D^{\text{pat}} \times N^{\text{pat}}}$, to query the pre-trained bi-LSTM word vectors $e$. For the new object-driven attention, we use the GloVe embedding of object class label $l_t^g$ to query the GloVe embedding of the words in the sentence, as illustrated in the lower part of Fig.~\ref{fig:attention}.

\noindent{\bf Feature map concatenation.}
The patch-wise context vector $c_j^{\text{pat}}$ can be directly concatenated with the image feature vector $h_j$ in the previous layer. However, the object-wise context vector $c_t^{\text{obj}}$ cannot, because they are associated with bounding boxes instead of pixels in the hidden feature map. We propose to copy the object-wise context vector $c_t^{\text{obj}}$ to every pixel where the $t$'th object is present, \ie, $M_t \otimes c_t^{\text{obj}}$ where $\otimes$ is the vector outer-product, as illustrated in the upper-right part of Fig.~\ref{fig:attention}. \footnote{This operation can be viewed as an inverse of the pooling operator.}

If there are multiple bounding boxes covering the same pixel, we have to decide whose context vector should be used on this pixel. In this case, we simply do a max-pooling across all the bounding boxes:
\begin{equation}\label{eqn:bthiden}
    c^{\text{obj}} = \max_{t~: 1\le t \le T} M_t \otimes c_{t}^{\text{obj}}.
\end{equation}
Then $c^{\text{obj}}$ can be concatenated with the feature map $h$ and patch-wise context vectors $c^{\text{pat}}$ for next-stage generation.

\noindent{\bf Label context vectors.}
Similarly, we distribute the class label information to the entire hidden feature map to get the label context vectors, \ie,
\begin{equation}\label{eqn:btlabel}
    c^{\text{lab}} = \max_{t~:~1\le t \le T} M_t \otimes e_{t}^{\text{g}}.
\end{equation}
Finally, we concatenate $h$, $c^{\text{pat}}$, $c^{\text{obj}}$ and $c^{\text{lab}}$ and pass the concatenated tensor through one up-sampling layer and several residual layers to generate a higher-resolution image.

\noindent{\bf Grid attention vs. object-driven attention.}
The process to compute the patch-wise context vectors above is the traditional grid attention mechanism used in AttnGAN~\cite{xu2017attngan}. Note that its attention weights $\beta_{j,i}^{\text{pat}}$ and context vector $c_j^{\text{pat}}$ are useful only when the hidden feature $h_j^{\text{pat}}$ in the $G_0$ stage correctly captures the content to be drawn in patch $j$. This essentially assumes that the generation in the $G_0$ stage already captures a rough sketch (semantic layout). This assumption is valid for simple datasets like birds~\cite{xu2017attngan}, but fails for complex datasets like COCO~\cite{LinMBHPRDZ14} where the generated low-resolution image $\hat{x}_0$ typically does {\it not} have a meaningful layout. In this case, the grid attention is even harmful, because patch-wise context vector is attended to a wrong word and thus generate the texture associated with that wrong word. This may be the reason why AttnGAN's generated image contains scattered patches of realistic texture but overall is semantically not meaningful; see Fig.~\ref{fig:maineg} for example. Similar phenomenon is also observed in DeepDream~\cite{mordvintsev2017deep}. On the contrary, in our object-driven attention, the attention weights $\beta_{t,i}^{\text{obj}}$ and context vector $c_t^{\text{obj}}$ rely on the class label $l_t^g$ of the bounding box and are independent of the generation in the $G_0$ stage. Therefore, the object-wise context vectors are always helpful to generate images that are consistent with the pre-generated semantic layout. Another benefit of this design is that the context vector $c_t^{\text{obj}}$ can also be used in the discriminator, as we present in $\S$~\ref{subsec:imageloss}.

\subsubsection{Discriminators}
\label{subsec:imageloss}
We design patch-wise and object-wise discriminators to train the attentive multi-stage generator above. Given a patch from uniformly-partitioned image patches determined by the uniform down-sampling structure of CNN, the patch-wise discriminator is trying to determine whether this patch is realistic or not (unconditional) and whether this patch is consistent with the sentence description or not (conditional). Given a bounding box and the class label of the object within it, the object-wise discriminator is trying to determine whether this region is realistic or not (unconditional) and whether this region is consistent with the sentence description and given class label or not (conditional).

\begin{figure}[tb]
\begin{center}
\includegraphics[width=1\linewidth]{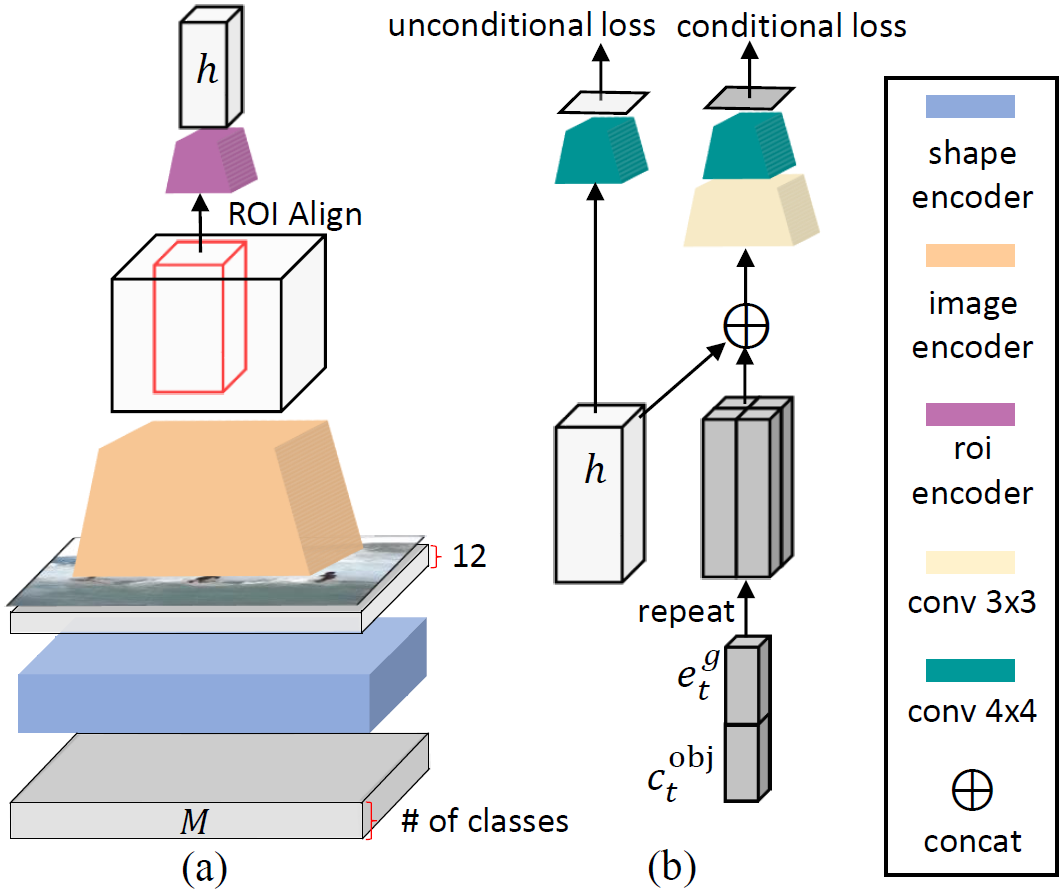}
\end{center}
\vspace{-13pt}
   \caption{\small Object-wise discriminator.}
	\label{fig:discriminator}
\vspace{-10pt}
\end{figure}

\noindent{\bf Patch-wise discriminators.} Given an image-sentence pair $x, \Bar{e}$ ($\Bar{e}$ is the sentence vector), the patch-wise unconditional and text discriminator can be written as
\begin{equation}\label{eq:patchD}
 \begin{aligned}
  p^{\text{pat,un}}= D_{\text{uncond.}}^{\text{pat}}(\text{Enc}(x)), \quad p^{\text{pat,con}} = D_{\text{text}}^{\text{pat}}(\text{Enc}(x), \Bar{e}),
 \end{aligned}
\end{equation}
where $\text{Enc}$ is a convolutional feature extractor that extracts patch-wise features, $D_{\text{uncond.}}$ ( $D_{\text{text}}^{\text{pat}}$ ) determine whether the patch is realistic (consistent with the text description) or not.

\noindent{\bf Shape discriminator.} In a similar manner, we have our patch-wise shape discriminator
\vspace{-2mm}
\begin{equation}\label{eq:patchshape}
 \begin{aligned}
  p^{\text{pix}} = D^{\text{pix}}(\text{Enc}(x, M)),
 \end{aligned}
\end{equation}
where we first concatenate the image $x$ and shapes $M$ in the channel dimension, and then extracts patch-wise features by another convolutional feature extractor $\text{Enc}$. The probabilities $p^{\text{pix}}$ determine whether the patch is consistent with the given shape. Our patch-wise discriminators $D_{\text{uncond.}}^{\text{pat}}$, $D_{\text{text}}^{\text{pat}}$ and $D^{\text{pix}}$ resembles the PatchGAN~\cite{pix2pix2017} for the image-to-image translation task. Compared with the global discriminators in AttnGAN~\cite{xu2017attngan}, the patch-wise discriminators not only reduce the model size and thus enable generating higher resolution images, but also increase the quality of generated images; see Table~\ref{tab:quantative} for experimental evidence.

\noindent{\bf Object-wise discriminators.} Given an image $x$, bounding boxes of objects $B_{1:T}$ and their shapes $M$, we propose the following object-wise discriminators:
\vspace{-2mm}
\begin{equation}\label{eq:instD}
 \begin{aligned}
  \{h_t^{\text{obj}}\}_{t=1}^T = &\text{FastRCNN}(x, M, B_{1:T}), \\
  p_t^{\text{obj,un}} =D_{\text{uncond.}}^{\text{obj}}(h_t^{\text{obj}}),& \quad p_t^{\text{obj,con}} = D^{\text{obj}}(h_t^{\text{obj}}, e_{t}^{\text{g}}, c_t^{\text{obj}}).
 \end{aligned}
\end{equation}
Here, we first concatenate the image $x$ and shapes $M$ and extract a region feature vector $h_t^{\text{obj}}$ for each bounding box through a Fast R-CNN model~\cite{girshick2015fast} with an ROI-align layer~\cite{he2017mask}; see Fig.~\ref{fig:discriminator}(a). Then similar to the patch-wise discriminator~\eqref{eq:patchD}, the unconditional (conditional) probabilities $p_t^{\text{obj,un}}$ ( $p_t^{\text{obj,con}}$) determine whether the $t$'th object is realistic (consistent with its class label $e_{t}^{\text{g}}$ and its text context information $c_t^{\text{obj}}$) or not; see Fig.~\ref{fig:discriminator}(b). Here, $e_{t}^{\text{g}}$ is the GloVe embedding of the class label and $c_t^{\text{obj}}$ is its text context information defined in \eqref{eq:AttnGAN}.

All discriminators are trained by the traditional cross entropy loss~\cite{goodfellow2014generative}.

\subsubsection{Loss function for the image generator}
The generator's GAN loss is a weighted sum of these discriminators' loss, \ie,
\begin{equation*}
% \scriptsize
\begin{aligned}
&\mathcal{L}_{\text{GAN}}(G) = - \frac{\lambda_{\text{obj}}}{T} \sum_{t=1}^{T} \left( \underbrace{\log p_t^{\text{obj,un}} }_\text{obj uncond. loss} + \underbrace{ \log p_t^{\text{obj,con}} }_\text{obj cond. loss} \right)\\
& - \frac{1}{N^{\text{pat}}} \sum_{j=1}^{N^{\text{pat}}} \left( \underbrace{\log p_j^{\text{pat,un}} }_\text{uncond. loss} + \underbrace{ \lambda_{\text{txt}} \log p_j^{\text{pat,con}} }_\text{text cond. loss} + \underbrace{ \lambda_{\text{pix}} \log p_j^{\text{pix}} }_\text{shape cond. loss}\right).
\end{aligned}
\end{equation*}
Here, $T$ is the number of bounding boxes, $N^{\text{pat}}$ is the number of regular patches, $(\lambda_{\text{obj}}, \lambda_{\text{txt}}, \lambda_{\text{pix}})$ are the weights of the object-wise GAN loss, patch-wise text conditional loss and patch-wise shape conditional loss, respectively. We tried combining our discriminators with the spectral normalized projection discriminator \cite{miyato2018spectral,miyato2018cgans}, but did not see significant performance improvement. We report performance of the spectral normalized version in $\S$~\ref{sec:exp:ablation} and provide model architecture details in $\S$~\ref{sec:appendix}.

Combined with the deep multi-modal attentive similarity model (DAMSM) loss introduced in \cite{xu2017attngan}, our final image generator's loss is
\begin{equation}\label{eqn:totalloss}
\begin{aligned}
    \mathcal{L}_G = \mathcal{L}_{\text{GAN}} + \lambda_{\text{DAMSM}} \mathcal{L}_{\text{DAMSM}}
\end{aligned}
\end{equation}
where $\lambda_{\text{damsm}}$ is a hyper-parameter to be tuned. Here, the DAMSM loss is a word level fine-grained image-text matching loss computed, which will be elaborated in $\S$~\ref{sec:appendix}. Based on the experiments on a held-out validation set, we set the hyperparameters in this section as: $\lambda_{\text{obj}} = 0.1, \lambda_{\text{txt}} = 0.1, \lambda_{\text{pix}} = 1$ and $\lambda_{\text{damsm}} = 100$.

\begin{remark}
Both the patch-wise and object-wise discriminators can be applied to different stages in the generation. We apply the patch-wise discriminator for every stage of the generation, following \cite{Han17stackgan2,pix2pix2017}, but only apply the object-wise discriminator at the final stage.
\end{remark}

\section{Experiments}
\label{sec:exp}
\noindent \textbf{Dataset.} We use the COCO dataset \cite{LinMBHPRDZ14} for evaluation. It contains 80 object classes, where each image is associated with object-wise annotations (\ie, bounding boxes and shapes) and 5 text descriptions. We use the official 2014 train (over 80K images) and validation (over 40K images) splits for training and test stages, respectively.

\begin{table}
\caption{\small The quantitative experiments. Methods marked with $0$, $1$ and $2$ respectively represent experiments using the predicted boxes and shapes, the ground-truth boxes and predicted shapes, and the ground-truth boxes and shapes. We use \textbf{bold}, $\ast$, and $\ast\ast$ to highlight the best performance under these three settings, respectively. The results of methods marked with $\dagger$ are those reported in the original papers. $\uparrow$ ($\downarrow$) means the higher (lower), the better.}
\centering
\scriptsize
\begin{tabular}[t]{|p{2.2cm}|c|c|c|}\hline
{Methods} &{Inception $\uparrow$} &{FID $\downarrow$} &{R-prcn ($\%$) $\uparrow$}\\ \hline\hline

{Obj-GAN$^{0}$} &$\mathbf{27.37 \pm 0.22}$ &$\mathbf{25.85}$ &$86.20 \pm 2.98$ \\ \hline
{Obj-GAN$^{1}$} &$27.96 \pm 0.39^{\ast}$ &$24.19^{\ast}$ &$88.36 \pm 2.82$ \\ \hline
{Obj-GAN$^{2}$} &$29.89 \pm 0.22^{\ast\ast}$ &$20.75^{\ast\ast}$ &$89.59 \pm 2.67$ \\ \hline

{P-AttnGAN w/ Lyt$^{0}$} &$18.84 \pm 0.29$ &$59.02$ &$65.71 \pm 3.74$ \\ \hline
{P-AttnGAN w/ Lyt$^{1}$} &$19.32 \pm 0.29$ &$54.96$ &$68.40 \pm 3.79$ \\ \hline
{P-AttnGAN w/ Lyt$^{2}$} &$20.81 \pm 0.16$ &$48.47$ &$70.94 \pm 3.70$ \\ \hline

{P-AttnGAN} &$26.31 \pm 0.43$ &$41.51$ &$86.71 \pm 2.97$ \\ \hline

{Obj-GAN w/ SN$^{0}$} &$26.97 \pm 0.31$  &$29.07$ &$\mathbf{86.84 \pm 2.82}$ \\ \hline
{Obj-GAN w/ SN$^{1}$} &$27.41 \pm 0.17$ &$27.26$ &$88.70 \pm 2.65^{\ast}$ \\ \hline
{Obj-GAN w/ SN$^{2}$} &$28.75 \pm 0.32$ &$23.37$ &$89.97 \pm 2.56^{\ast\ast}$ \\ \hline \hline

{Reed \etal \cite{reed2016generative}$\dagger$} &$7.88 \pm 0.07$  &n/a &n/a \\ \hline
{StackGAN \cite{Han16stackgan}$\dagger$} &$8.45 \pm 0.03$ &n/a &n/a  \\ \hline
{AttnGAN \cite{xu2017attngan}} &$23.79 \pm 0.32$ &$28.76$ &$82.98 \pm 3.15$ \\ \hline
{vmGAN \cite{zhang18vmgan}$\dagger$} &$9.94 \pm 0.12$ &n/a &n/a  \\ \hline
{Sg2Im \cite{johnson2018image}$\dagger$} &$6.7 \pm 0.1$ &n/a &n/a  \\ \hline

{Infer \cite{hong2018inferring}$^{0}\dagger$} &$11.46 \pm 0.09$ &n/a &n/a \\ \hline
{Infer \cite{hong2018inferring}$^{1}\dagger$} &$11.94 \pm 0.09$ &n/a &n/a \\ \hline
{Infer \cite{hong2018inferring}$^{2}\dagger$} &$12.40 \pm 0.08$ &n/a &n/a \\ \hline

{Obj-GAN-SOTA$^{0}$} &$30.29 \pm 0.33$ &$25.64$ &$91.05 \pm 2.34$ \\ \hline
{Obj-GAN-SOTA$^{1}$} &$30.91 \pm 0.29$ &$24.28$ &$92.54 \pm 2.16$ \\ \hline
{Obj-GAN-SOTA$^{2}$} &$32.79 \pm 0.21$ &$21.21$ &$93.39 \pm 2.08$ \\ \hline
\end{tabular}
\label{tab:quantative}
\vspace{-13pt}
\end{table}

\noindent \textbf{Evaluation metrics.} We use the Inception score~\cite{Salimans2016} and \textit{Fr\'echet inception distance} (FID)~\cite{heusel2017gans} score as the quantitative evaluation metrics. In our experiments, we found that Inception score can be saturated, even over-fitted, while FID is a more robust measure and aligns better with human qualitative evaluation. Following~\cite{xu2017attngan}, we also use R-precision, a common evaluation metric for ranking retrieval results, to evaluate whether the generated image is well conditioned on the given text description. More specifically, given a pre-trained image-to-text retrieval model, we use generated images to query their corresponding text descriptions. First, given generated image $\hat{x}$ conditioned on sentence $s$ and 99 random sampled sentences $\{s'_i: 1\le i \le 99\}$, we rank these 100 sentences by the pre-trained image-to-text retrieval model. If the ground truth sentence $s$ is ranked highest, we count this a success retrieval. For all the images in the test dataset, we perform this retrieval task once and finally count the percentage of success retrievals as the R-precision score.

It is important to point out that none of these quantitative metrics are perfect. Better metrics are required to evaluate image generation qualities in complicated scenes. In fact, the Inception score completely fails in evaluating the semantic layout of the generated images. The R-precision score depends on the pre-trained image-to-text retrieval model it uses, and can only capture the aspects that the retrieval model is able to capture. The pre-trained model we use is still limited in capturing the relations between objects in complicated scenes, so is our R-precision score.

\noindent \textbf{Quantitative evaluation.} We compute these three metrics under two settings for the full validation dataset.

\noindent \textbf{Qualitative evaluation.} Apart from the quantitative evaluation, we also visualize the outputs of all ablative versions of Obj-GAN and the state-of-the-art methods (\ie, \cite{xu2017attngan}) whose pre-trained models are publicly available.

\begin{figure*}[tb]
\begin{center}
\includegraphics[width=1\linewidth]{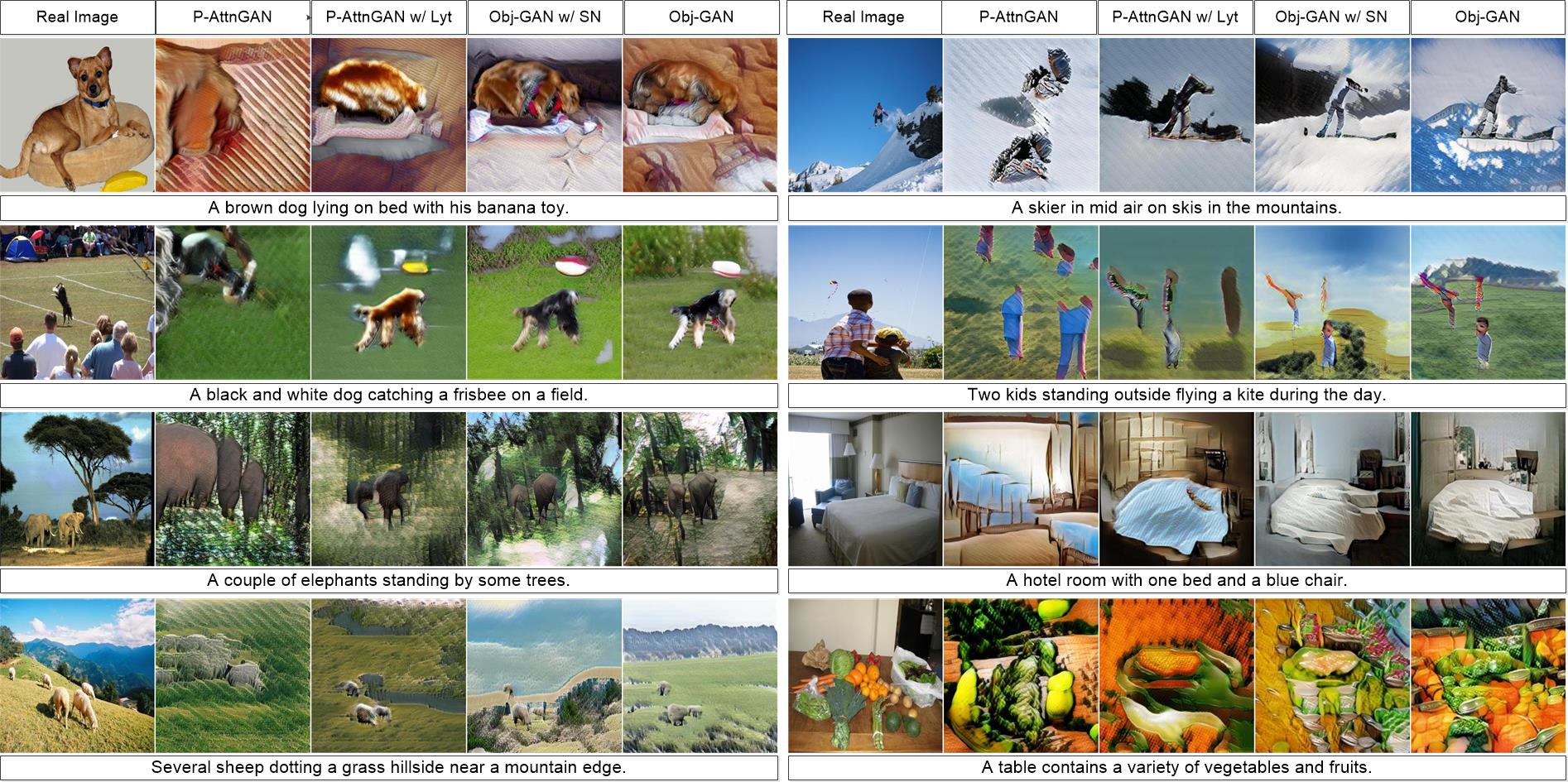}
\end{center}
\vspace{-18pt}
   \caption{\small The overall qualitative comparison. All images are generated without the usage of any ground-truth information.}
	\label{fig:overall_qualitative_comparison}
\vspace{-17pt}
\end{figure*}

\subsection{Ablation study}
\label{sec:exp:ablation}
In this section, we first evaluate the effectiveness of the object-driven attention. Next, we compare the object-driven attention mechanism with the grid attention mechanism. Then, we evaluate the impact of the spectral normalization for Obj-GAN. We use Fig.~\ref{fig:overall_qualitative_comparison} and the higher half of Table~\ref{tab:quantative} to present the comparison among different ablative versions of Obj-GAN. Note that all ablative versions have been trained with batch size $16$ for $60$ epochs. In addition, we use the lower half of Table~\ref{tab:quantative} to show the comparison between Obj-GAN and previous methods. Finally, we validated the Obj-GAN's generalization ability on the novel text descriptions.

\noindent \textbf{Object-driven attention.} To evaluate the efficacy of the object-driven attention mechanism, we implement a baseline, named P-AttnGAN w/ Lyt, by disabling the object-driven attention mechanism in Obj-GAN. In essence, P-AttnGAN w/ Lyt can be considered as an improved version of AttnGAN with the patch-wise discriminator (abbreviated as the prefix ``P-" in name) and the modules (\eg, shape discriminator) for handling the conditional layout (abbreviated as ``Lyt"). Moreover, it can also be considered as a modified implementation of \cite{hong2018inferring}, which resembles their two-step (layout-image) generation. Note that there are three key differences between P-AttnGAN w/ Lyt and \cite{hong2018inferring}: (i) P-AttnGAN w/ Lyt has a multi-stage image generator that gradually increases the generated resolution and refines the generated images, while \cite{hong2018inferring} has a single-stage image generator. (ii) With the help of the grid attentive module, P-AttnGAN w/ Lyt is able to utilize the fine-grained word-level information, while \cite{hong2018inferring} conditions on the global sentence information. (iii) The third difference lies in their loss functions: P-AttnGAN w/ Lyt uses the DAMSM loss in \eqref{eqn:totalloss} to penalize the mismatch between the generated images and the input text descriptions, while \cite{hong2018inferring} uses the perceptual loss to penalize the mismatch between the generated images and the ground-truth images. As shown in Table~\ref{tab:quantative}, P-AttnGAN w/ Lyt yields higher Inception score than \cite{hong2018inferring} does.

\begin{figure}[tb]
\begin{center}
\includegraphics[width=1\linewidth]{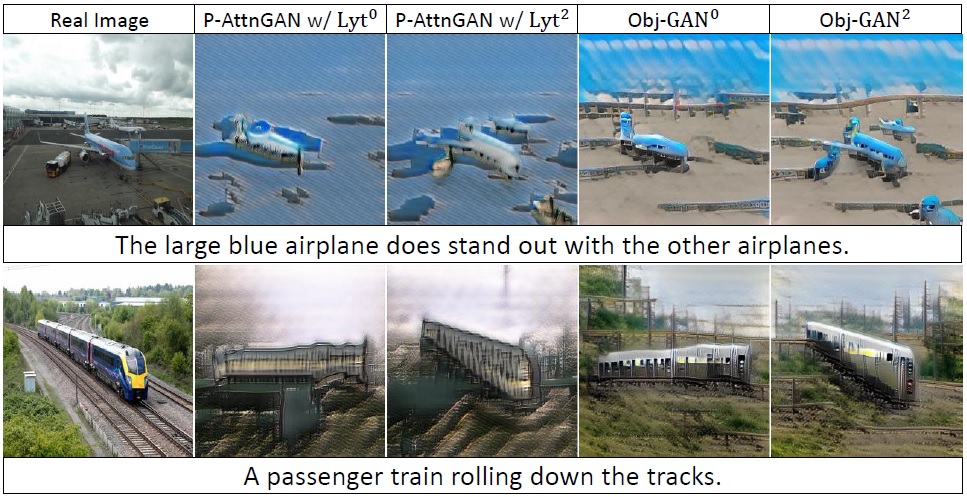}
\end{center}
\vspace{-16pt}
   \caption{\small Qualitative comparison with P-AttnGAN w/ Lyt.}
	\label{fig:comp_w_honglak}
\vspace{-13pt}
\end{figure}

We compare Obj-GAN with P-AttnGAN w/ Lyt under three settings, with each corresponding to a set of conditional layout input, \ie, the predicted boxes $\&$ shapes, the ground-truth boxes $\&$ predicted boxes, and the ground-truth boxes $\&$ shapes. As presented in Table~\ref{tab:quantative}, Obj-GAN consistently outperforms P-AttnGAN w/ Lyt on all three metrics. In Fig.~\ref{fig:comp_w_honglak}, we use the same layout as the conditional input, and compare the visual quality of their generated images. An interesting phenomenon shown in Fig.~\ref{fig:comp_w_honglak} is that both the foreground objects (\eg, airplane and train) and the background (\eg, airport and trees) textures synthesized by Obj-GAN are much richer and smoother than those using P-AttnGAN w/ Lyt. The effectiveness of the object-driven attention for the foreground objects is easy to understand. The benefits for the background textures using the object-driven attention mechanism is probably due to the fact that it implicitly provides stronger signal that distinguishes the foreground. As such, the image generator may have richer guidance and clearer emphasis when synthesizing textures for a certain region.

\begin{figure}[tb]
\begin{center}
\includegraphics[width=1\linewidth]{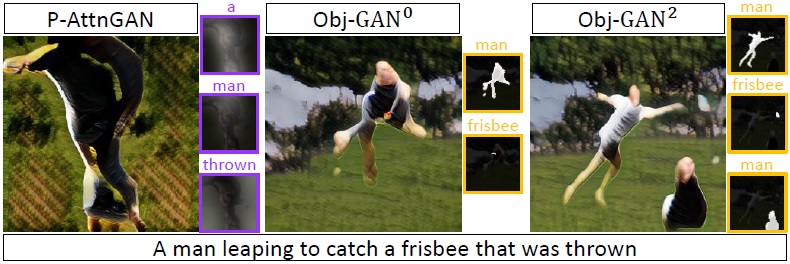}
\end{center}
\vspace{-16pt}
   \caption{\small Qualitative comparison with P-AttnGAN. The attention maps of each method are shown beside the generated image.}
	\label{fig:comp_w_attngan}
\vspace{-13pt}
\end{figure}

\noindent \textbf{Grid attention vs. object-driven attention.} We compare Obj-GAN with P-AttnGAN herein, so as to compare the effects of the object-driven and the grid attention mechanisms. In Fig.~\ref{fig:comp_w_attngan}, we show the generated image of each method as well as the corresponding attention maps aligned on the right side. In a grid attention map, the brightness of a region reflects how much this region attended to the word above the map. As for the object-driven attention map, the word above each attention map is the most attended word by the highlighted object. The highlighted region of an object-driven attention map is the object shape.

As analyzed in $\S$~\ref{subsec:imagegenerator}, the reliability of grid attention weights depends on the quality of the previous layer's image region features. This makes the grid attention unreliable sometimes, especially for complex scenes. For example, the grid attention weights in Fig.~\ref{fig:comp_w_attngan} are unreliable because they are scattered (\eg, the attention map for ``man") and inaccurate. However, this is not a problem for the object-driven attention mechanism, because its attention weights are directly calculated from embedding vectors of words and class labels and are independent of image features. Moreover, as shown in Fig.~\ref{fig:attention} and Equ.~\eqref{eqn:bthiden}, the impact region of the object-driven attention context vector is bounded by the object shapes, which further enhances its semantics meaningfulness. As a result, the instance-driven attention significantly improves the visual quality of the generated images, as demonstrated in Fig.~\ref{fig:comp_w_attngan}. Moreover, the performance can be further improved if the semantic layout generation is improved. In the extreme case, Obj-GAN based on ground truth layout (Obj-GAN$^{2}$) has the best visual quality (the rightmost column of Fig.~\ref{fig:comp_w_attngan}) and the best quantitative evaluation (Table~\ref{tab:quantative}).

\noindent \textbf{Obj-GAN w/ SN vs. Obj-GAN.}
We present the comparison between the cases with or without spectral normalization in the discriminators in Table~\ref{tab:quantative} and Fig.~\ref{fig:overall_qualitative_comparison}. We observe that there is no obvious improvement on the visual quality, but slightly worse on the quantitative metrics. We show more results and discussions in $\S$~\ref{sec:appendix}.

\noindent \textbf{Comparison with previous methods.}
To compare Obj-GAN with the previous methods, initialized by the Obj-GAN models in the ablation study, we trained Obj-GAN-SOTA with batch size $64$ for 10 more epochs.
%All results except AttnGAN \cite{xu2017attngan} are quoted from the previous papers \cite{hong2018inferring,johnson2018image,zhang18vmgan}.
In order to evaluate AttnGAN on FID, we conducted the evaluation on the officially released pre-trained model. Note that the Sg2Im \cite{johnson2018image} focuses on generating images from scene graphs and conducted the evaluation on a different split of COCO. However, we still included Sg2Im's results to reflect the broader context of the related topic. As shown in Table~\ref{tab:quantative}, Obj-GAN-SOTA outperforms all previous methods significantly. We notice that the increment of batch size does boost the Inception score and R-precision, but does not improve FID. The possible explanation is: with a larger batch size, the DAMSM loss (a ranking loss in essence) in \eqref{eqn:totalloss} plays a more important role and improves Inception and R-precision, but it does not focus on reducing FID between the generated images and the real ones.

\begin{figure}[tb]
\begin{center}
\includegraphics[width=1\linewidth]{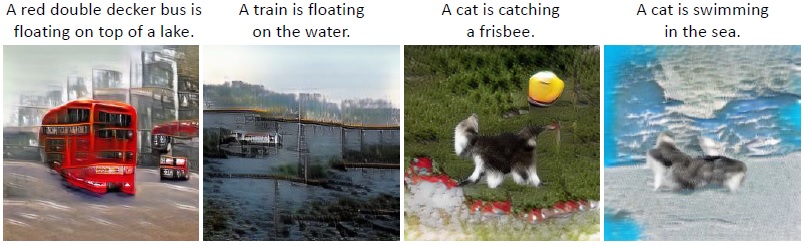}
\end{center}
\vspace{-13pt}
   \caption{\small Generated images for novel descriptions.}
	\label{fig:generalization}
	\vspace{-13pt}
\end{figure}

\noindent \textbf{Generalization ability.}
We further investigate if Obj-GAN just memorizes the scenarios in COCO or it indeed learns the relations between the objects and their surroundings. To this end, we compose several descriptions which reflect novel scenarios that are unlikely to happen in the real-world, \eg, \emph{a decker bus is floating on top of a lake}, or \emph{a cat is catching a frisbee}. We use Obj-GAN to synthesize images for these rare scenes. The results in Fig.~\ref{fig:generalization} further demonstrate the good generalization ability of Obj-GAN.

\section{Conclusions}
\label{sec:conclusion}
In this paper, we have presented a multi-stage Object-driven Attentive Generative Adversarial Networks (Obj-GANs) for synthesizing images with complex scenes from the text descriptions. With a novel object-driven attention layer at each stage, our generators are able to utilize the fine-grained word/object-level information to gradually refine the synthesized image. We also proposed the Fast R-CNN based object-wise discriminators, each of which is paired with a conditional input of the generator and provides object-wise discrimination signal for that condition. Our Obj-GAN significantly outperforms previous state-of-the-art GAN models on various metrics on the large-scale challenging COCO benchmark. Extensive experiments demonstrate the effectiveness and generalization ability of Obj-GAN on text-to-image generation for complex scenes.

{\small
\bibliographystyle{ieee}
\bibliography{paperbib}
}

\appendix

\section{Appendix}
\label{sec:appendix}
% the \\ insures the section title is centered below the phrase: AppendixA

\begin{figure*}[tb]
\begin{center}
\includegraphics[width=1\linewidth]{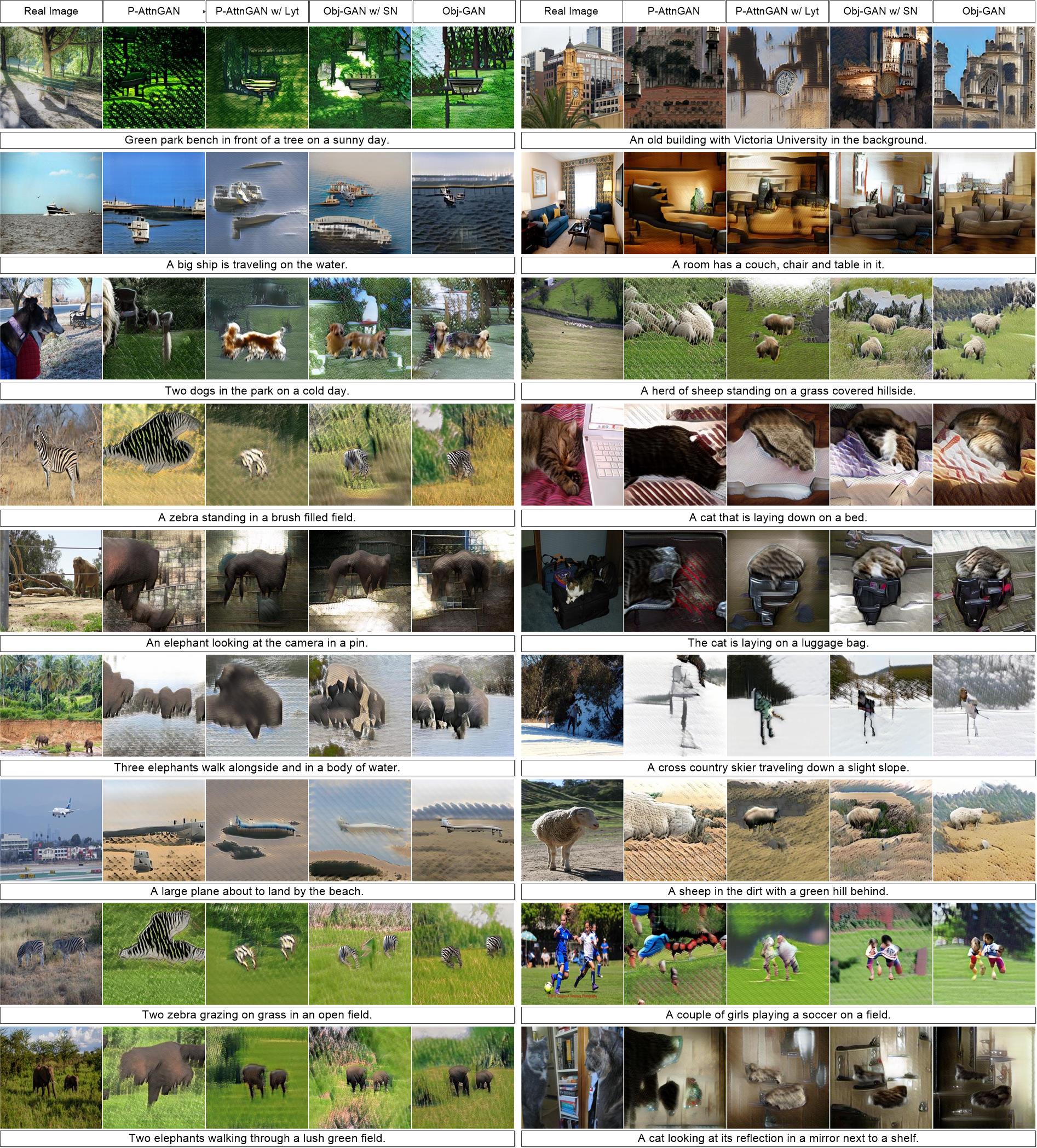}
\vspace{-0.5cm}
\caption{\small The overall qualitative comparison.}
\label{fig:more_overall_qualitative_comparison1}
\vspace{-0.7cm}
\end{center}
\end{figure*}

\begin{figure*}[tb]
\begin{center}
\includegraphics[width=1\linewidth]{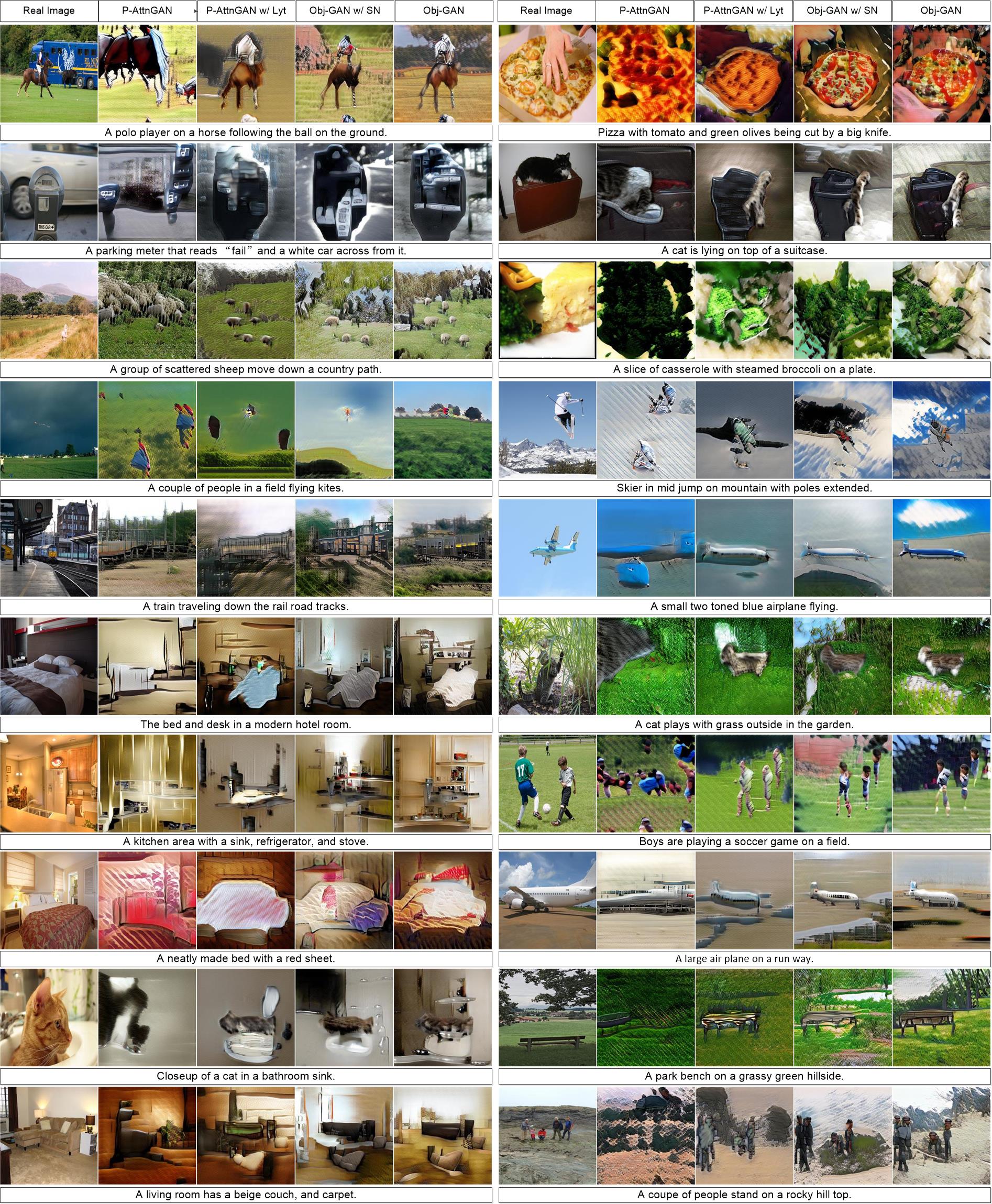}
\vspace{-0.3cm}
\caption{\small The overall qualitative comparison.}
\label{fig:more_overall_qualitative_comparison2}
\vspace{-0.5cm}
\end{center}
\end{figure*}

\subsection{Obj-GAN vs. the ablative versions}
\label{sec:appendix:ablation}
In this section, we show more images generated by our Obj-GAN and its ablative versions on the COCO dataset. In Fig.~\ref{fig:more_overall_qualitative_comparison1} and Fig.~\ref{fig:more_overall_qualitative_comparison2}, we provide more comparisons as the complementary for Fig.~\ref{fig:overall_qualitative_comparison}. It can be found that there are no obvious improvement on the visual quality when using the spectral normalization.

\subsection{Visualization of attention maps}
\label{sec:appendix:attnmaps}
\begin{figure*}[h]
\begin{center}
\vspace{-0.3cm}
\includegraphics[width=1\linewidth]{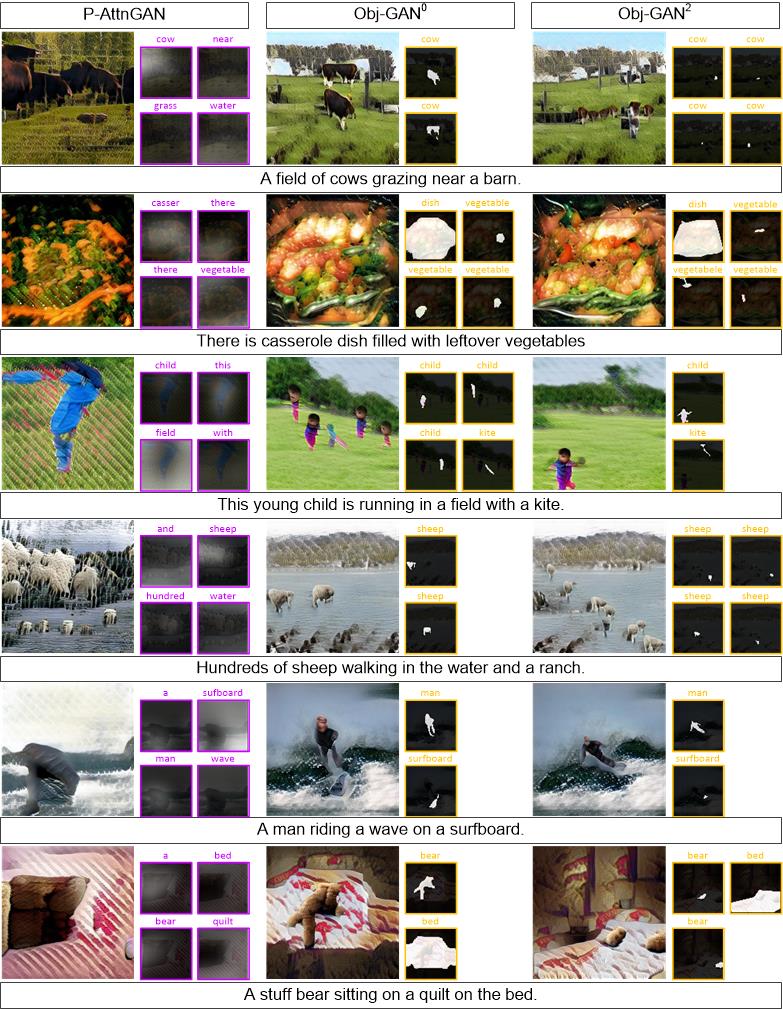}
\caption{\small Qualitative comparison with P-AttnGAN. The attention maps of each method are shown beside the generated image.}
\label{fig:more_comp_w_attngan}
\vspace{-0.3cm}
\end{center}
\end{figure*}

In Fig.~\ref{fig:more_comp_w_attngan}, we visualize attention maps generated by P-AttnGAN and Obj-GAN as the complementary for Fig.~\ref{fig:comp_w_attngan}.

\subsection{Results based on the ground-truth layout}
\label{sec:appendix:gt}
\begin{figure*}[h]
        \centering
        \begin{subfigure}[b]{0.23\textwidth}
                \includegraphics[width=\textwidth]{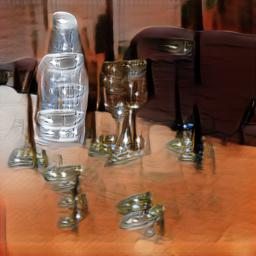}
                \caption{A glass table with a bottle and glass of wine next to a chair.}
        \end{subfigure}
        ~
        \begin{subfigure}[b]{0.23\textwidth}
                \includegraphics[width=\textwidth]{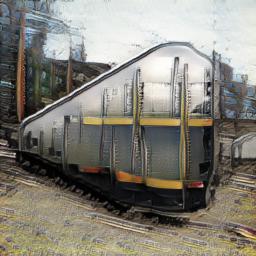}
                \caption{A train sitting on some tracks next to a sidewalk.}
        \end{subfigure}
        ~
        \begin{subfigure}[b]{0.23\textwidth}
                \includegraphics[width=\textwidth]{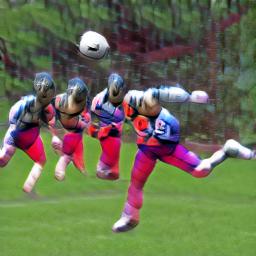}
                \caption{Soccer player wearing green and orange hitting soccer ball.}
        \end{subfigure}
        ~
        \begin{subfigure}[b]{0.23\textwidth}
                \includegraphics[width=\textwidth]{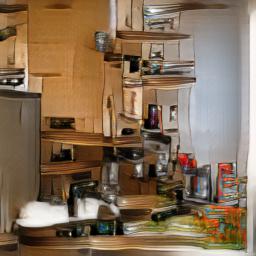}
                \caption{A kitchen with a very messy counter space.}
        \end{subfigure}
        ~
        \begin{subfigure}[b]{0.23\textwidth}
                \includegraphics[width=\textwidth]{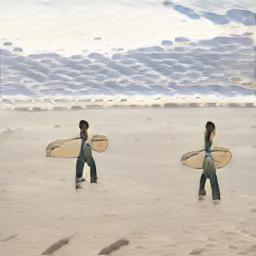}
                \caption{The people are on the beach getting ready to surf.}
        \end{subfigure}
        ~
        \begin{subfigure}[b]{0.23\textwidth}
                \includegraphics[width=\textwidth]{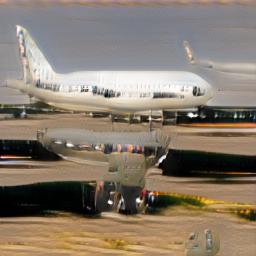}
                \caption{A jet airliner waits its turn on the runway.}
        \end{subfigure}
        ~
        \begin{subfigure}[b]{0.23\textwidth}
                \includegraphics[width=\textwidth]{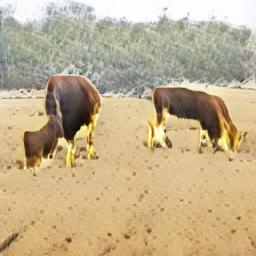}
                \caption{Two cows are grazing in a dirt field.}
        \end{subfigure}
        ~
        \begin{subfigure}[b]{0.23\textwidth}
                \includegraphics[width=\textwidth]{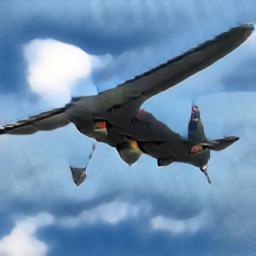}
                \caption{A small lightweight airplane flying through the sky.}
        \end{subfigure}
        ~
        \begin{subfigure}[b]{0.23\textwidth}
                \includegraphics[width=\textwidth]{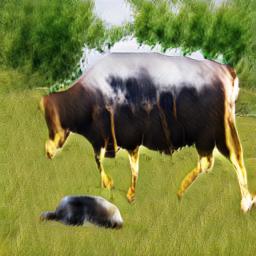}
                \caption{A cow running in a field next to a dog.}
        \end{subfigure}
        ~
        \begin{subfigure}[b]{0.23\textwidth}
                \includegraphics[width=\textwidth]{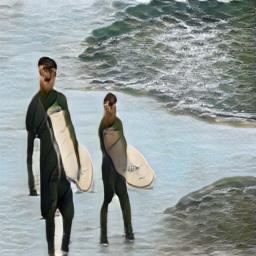}
                \caption{Two people go into the water with their surfboards.}
        \end{subfigure}
        ~
        \begin{subfigure}[b]{0.23\textwidth}
                \includegraphics[width=\textwidth]{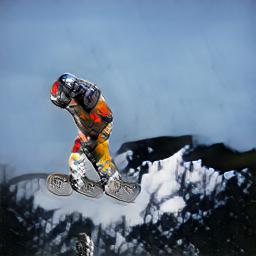}
                \caption{A man in a helmet jumps a snowboard.}
        \end{subfigure}
        ~
        \begin{subfigure}[b]{0.23\textwidth}
                \includegraphics[width=\textwidth]{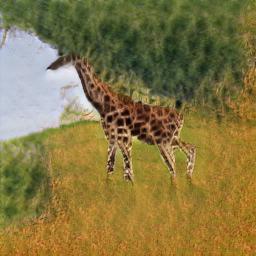}
                \caption{A giraffe is standing all alone in a grassy area.}
        \end{subfigure}
        ~
        \begin{subfigure}[b]{0.23\textwidth}
                \includegraphics[width=\textwidth]{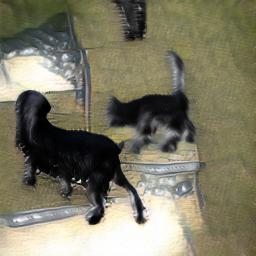}
                \caption{The black dog is staring at the cat.}
        \end{subfigure}
        ~
        \begin{subfigure}[b]{0.23\textwidth}
                \includegraphics[width=\textwidth]{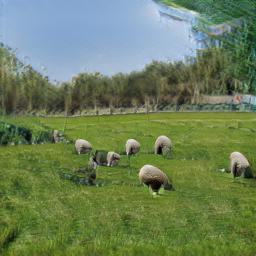}
                \caption{A bunch of sheep are standing in a field.}
        \end{subfigure}
        ~
        \begin{subfigure}[b]{0.23\textwidth}
                \includegraphics[width=\textwidth]{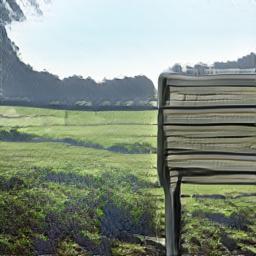}
                \caption{A bench sitting on top of  a lush green hillside.}
        \end{subfigure}
        ~
        \begin{subfigure}[b]{0.23\textwidth}
                \includegraphics[width=\textwidth]{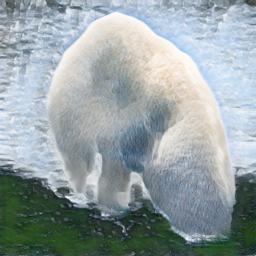}
                \caption{A polar bear playing in the water at a wild life enclosure.}
        \end{subfigure}
\caption{\small Results based on the ground-truth layout.}
\label{fig:gt_results1}
\vspace{-0.3cm}
\end{figure*}

\begin{figure*}[h]
        \centering
        \begin{subfigure}[b]{0.23\textwidth}
                \includegraphics[width=\textwidth]{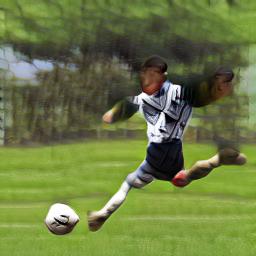}
                \caption{A man on a soccer field next to a ball.}
        \end{subfigure}
        ~
        \begin{subfigure}[b]{0.23\textwidth}
                \includegraphics[width=\textwidth]{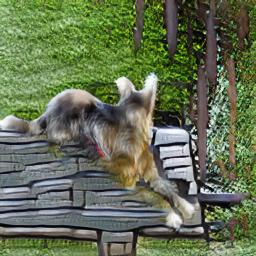}
                \caption{A dog sitting on a bench in front of a garden.}
        \end{subfigure}
        ~
        \begin{subfigure}[b]{0.23\textwidth}
                \includegraphics[width=\textwidth]{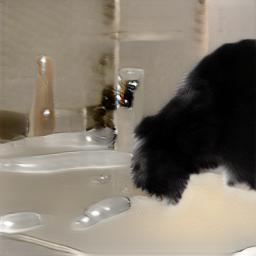}
                \caption{A black cat drinking water out of a water faucet.}
        \end{subfigure}
        ~
        \begin{subfigure}[b]{0.23\textwidth}
                \includegraphics[width=\textwidth]{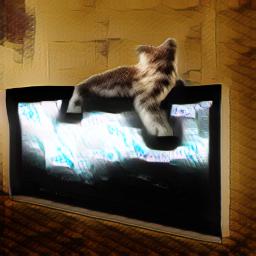}
                \caption{A cat laying on a TV in the middle of the room.}
        \end{subfigure}
        ~
        \begin{subfigure}[b]{0.23\textwidth}
                \includegraphics[width=\textwidth]{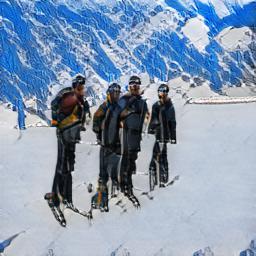}
                \caption{Four people on skis below a mountain taking a picture.}
        \end{subfigure}
        ~
        \begin{subfigure}[b]{0.23\textwidth}
                \includegraphics[width=\textwidth]{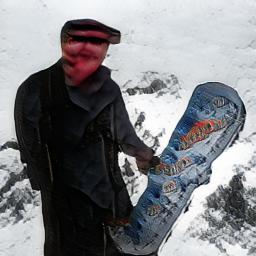}
                \caption{A man in outdoor winter clothes holds a snowboard.}
        \end{subfigure}
        ~
        \begin{subfigure}[b]{0.23\textwidth}
                \includegraphics[width=\textwidth]{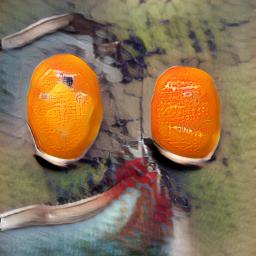}
                \caption{A orange before and after it was cu.}
        \end{subfigure}
        ~
        \begin{subfigure}[b]{0.23\textwidth}
                \includegraphics[width=\textwidth]{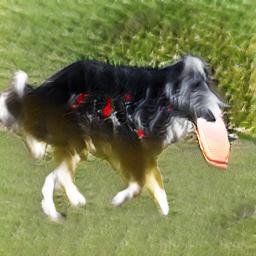}
                \caption{A dog running with a frisbee in its mouth.}
        \end{subfigure}
        ~
        \begin{subfigure}[b]{0.23\textwidth}
                \includegraphics[width=\textwidth]{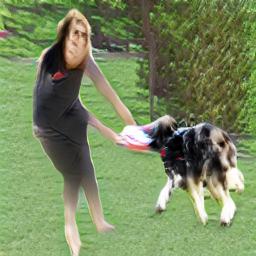}
                \caption{A woman and a dog tussle over a frisbee.}
        \end{subfigure}
        ~
        \begin{subfigure}[b]{0.23\textwidth}
                \includegraphics[width=\textwidth]{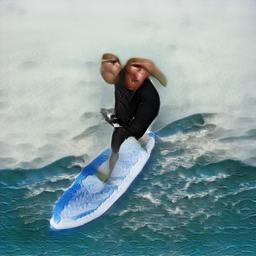}
                \caption{Man in a wetsuit on top of a blue and white surfboard.}
        \end{subfigure}
        ~
        \begin{subfigure}[b]{0.23\textwidth}
                \includegraphics[width=\textwidth]{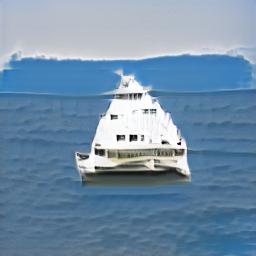}
                \caption{A white ship sails in the blue ocean water.}
        \end{subfigure}
        ~
        \begin{subfigure}[b]{0.23\textwidth}
                \includegraphics[width=\textwidth]{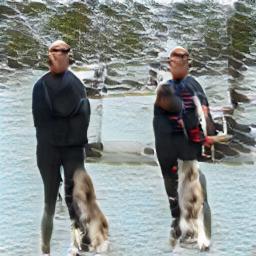}
                \caption{A couple of men standing next to dogs near water.}
        \end{subfigure}
        ~
        \begin{subfigure}[b]{0.23\textwidth}
                \includegraphics[width=\textwidth]{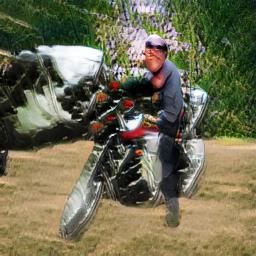}
                \caption{A man on a motorcycle in a carport.}
        \end{subfigure}
        ~
        \begin{subfigure}[b]{0.23\textwidth}
                \includegraphics[width=\textwidth]{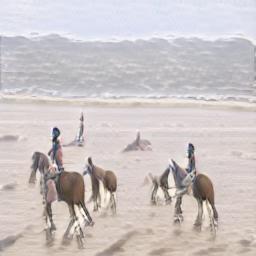}
                \caption{A group of people riding horses on a beach.}
        \end{subfigure}
        ~
        \begin{subfigure}[b]{0.23\textwidth}
                \includegraphics[width=\textwidth]{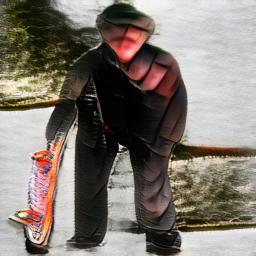}
                \caption{A hipster wearing flood pants poses with his skateboard.}
        \end{subfigure}
        ~
        \begin{subfigure}[b]{0.23\textwidth}
                \includegraphics[width=\textwidth]{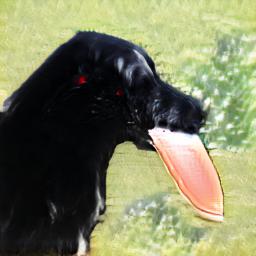}
                \caption{A black dog holding a frisbee in its mouth.}
        \end{subfigure}
\caption{\small Results based on the ground-truth layout.}
\label{fig:gt_results2}
\vspace{-0.3cm}
\end{figure*}

\begin{figure*}[h]
        \centering
        \begin{subfigure}[b]{0.23\textwidth}
                \includegraphics[width=\textwidth]{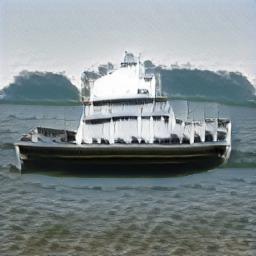}
                \caption{A big boat on the water near the shore.}
        \end{subfigure}
        ~
        \begin{subfigure}[b]{0.23\textwidth}
                \includegraphics[width=\textwidth]{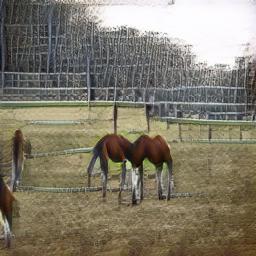}
                \caption{All the horses in the pen are grazing.}
        \end{subfigure}
        ~
        \begin{subfigure}[b]{0.23\textwidth}
                \includegraphics[width=\textwidth]{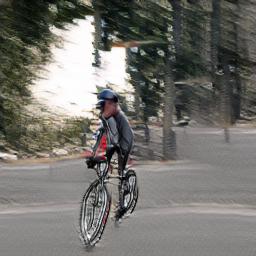}
                \caption{A man riding a bike down the middle of a street.}
        \end{subfigure}
        ~
        \begin{subfigure}[b]{0.23\textwidth}
                \includegraphics[width=\textwidth]{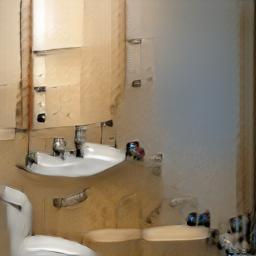}
                \caption{A bathroom with a sink and a toilet.}
        \end{subfigure}
        ~
        \begin{subfigure}[b]{0.23\textwidth}
                \includegraphics[width=\textwidth]{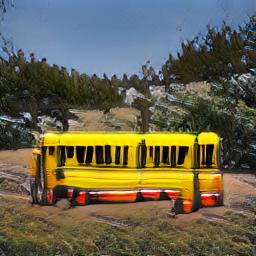}
                \caption{A yellow school bus parked  near a tree.}
        \end{subfigure}
        ~
        \begin{subfigure}[b]{0.23\textwidth}
                \includegraphics[width=\textwidth]{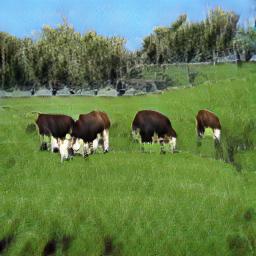}
                \caption{A group of cows graze on some grass.}
        \end{subfigure}
        ~
        \begin{subfigure}[b]{0.23\textwidth}
                \includegraphics[width=\textwidth]{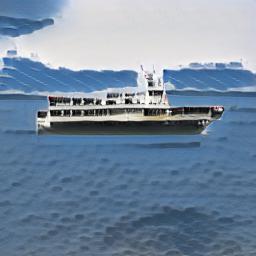}
                \caption{A ship is sailing across an ocean filled with waves.}
        \end{subfigure}
        ~
        \begin{subfigure}[b]{0.23\textwidth}
                \includegraphics[width=\textwidth]{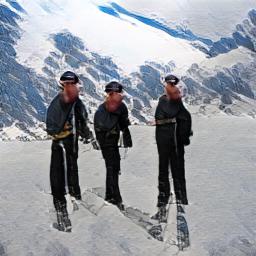}
                \caption{Three skiers posing for a picture on the slope.}
        \end{subfigure}
        ~
        \begin{subfigure}[b]{0.23\textwidth}
                \includegraphics[width=\textwidth]{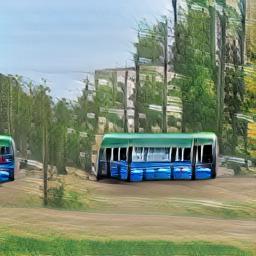}
                \caption{A large green bus approaching a bus stop.}
        \end{subfigure}
        ~
        \begin{subfigure}[b]{0.23\textwidth}
                \includegraphics[width=\textwidth]{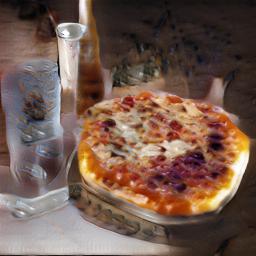}
                \caption{A close view of a pizza, and a mug of beer.}
        \end{subfigure}
        ~
        \begin{subfigure}[b]{0.23\textwidth}
                \includegraphics[width=\textwidth]{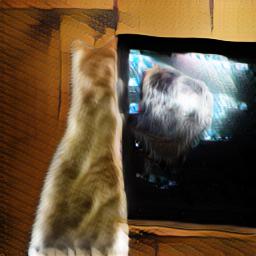}
                \caption{A cat is looking at a television displaying a dog in a cage.}
        \end{subfigure}
        ~
        \begin{subfigure}[b]{0.23\textwidth}
                \includegraphics[width=\textwidth]{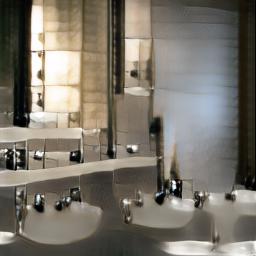}
                \caption{Three white sinks in a bathroom under mirrors.}
        \end{subfigure}
        ~
        \begin{subfigure}[b]{0.23\textwidth}
                \includegraphics[width=\textwidth]{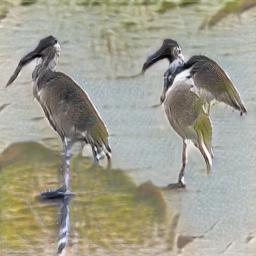}
                \caption{Three cranes standing on one leg in the water.}
        \end{subfigure}
        ~
        \begin{subfigure}[b]{0.23\textwidth}
                \includegraphics[width=\textwidth]{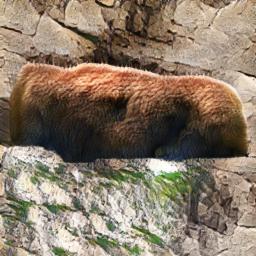}
                \caption{A bear lying on a rock in its den, looking upward.}
        \end{subfigure}
        ~
        \begin{subfigure}[b]{0.23\textwidth}
                \includegraphics[width=\textwidth]{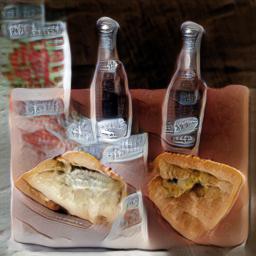}
                \caption{Two bottles of soda sit near a sandwich.}
        \end{subfigure}
        ~
        \begin{subfigure}[b]{0.23\textwidth}
                \includegraphics[width=\textwidth]{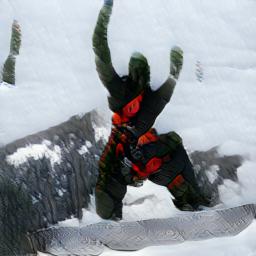}
                \caption{Someone on a snowboard coming to a stop.}
        \end{subfigure}
\caption{\small Results based on the ground-truth layout.}
\label{fig:gt_results3}
\vspace{-0.3cm}
\end{figure*}

We show the results generated by Obj-GAN based on the ground-truth layout in Fig.~\ref{fig:gt_results1}, Fig.~\ref{fig:gt_results2} and Fig.~\ref{fig:gt_results3}.

\clearpage

\subsection{Bi-LSTM text encoder, DAMSM and R-precision}
\label{sec:DAMSM}

We use the deep attentive multi-modal similarity model (DAMSM) proposed in \cite{xu2017attngan}, which learns a joint embedding of the image regions and words of a sentence in a common semantic space. The fine-grained conditional loss enforces the sub-region of the generated image to match the corresponding word in the sentence.

\textbf{Bi-LSTM text encoder} serves as the text encoder for both DAMSM and the box generator (see $\S$~\ref{sec:layout}). Bi-LSTM text encoder is a bi-directional LSTM that extracts semantic vectors from the text description. In the Bi-LSTM, each word corresponds to two hidden states, one for each direction. Thus, we concatenate its two hidden states to represent the semantic meaning of a word. The feature matrix of all words is indicated by $\dot{e} \in \mathbb{R}^{D\times T_s}$. Its $i^{th}$ column $\dot{e}_i$ is the feature vector for the $i^{th}$ word. $D$ is the dimension of the word vector and $T_s$ is the number of words. Meanwhile, the last hidden states of the bi-directional LSTM are concatenated to be the global sentence vector, denoted by $\hat{e} \in \mathbb{R}^{D}$. We present the network architectures for the Bi-LSTM text encoder in Table~\ref{tab:bilstm}.

\textbf{The image encoder }{
%Our image encoder
is a convolutional neural network that maps images to semantic vectors.
The intermediate layers of the CNN model learns local features of different regions of the image, while the later layers learn global features of the image.
More specifically, the image encoder is built upon Inception-v3 model pre-trained on ImageNet.
We first rescale the input image to be 299$\times$299 pixels.
And then, we extract the local feature matrix $f \in \mathbb{R}^{768\times289}$ (reshaped from 768$\times$17$\times$17) from ``$mixed\_6e$'' layer of Inception-v3.
Each column of $f$ is the feature vector of a local image region.
768 is the dimension of the local feature vector, and 289 is the number of regions in the image.
Meanwhile, the global feature vector $\overline{f} \in \mathbb{R}^{2048}$ is extracted from the last average pooling layer of Inception-v3.
Finally, we convert the image features to the common semantic space of text features by adding a new layer perceptron as shown in Eq.~(\ref{eq:localFtr}),
 \begin{equation}\label{eq:localFtr}
  v = Wf; \, \, \, \,\, \, \, \, \overline{v} = \overline{W}\,\overline{f},
 \end{equation}
where $v \in \mathbb{R}^{D\times289}$ and its $i^{th}$ column $v_{i}$ is the visual
feature vector for the $i^{th}$ image region; $\overline{v} \in \mathbb{R}^{D}$ is the visual feature vector for the whole image.
While $v_{i}$ is the local image feature vector that corresponds to the word embedding, $\overline{v}$ is the global feature vector that is related to the sentence embedding. $D$ is the dimension of the multimodal (\ie, image and text modalities) feature space.
For efficiency, all parameters in layers built from Inception-v3 model are fixed, and the parameters in newly added layers are jointly learned with the rest of networks.
}

\textbf{The fine-grained conditional loss }{
is designed to learn the correspondence between image regions and words.
However, it is difficult to obtain manual annotations.
% It is difficult to manually annotate the correspondence between image regions and words.
Actually, many words relate to concepts that may not easily be visually defined, such as \textit{open} or \textit{old}.
One possible solution is to learn word-image correspondence in a semi-supervised manner, in which the only supervision is the correspondence between the entire image and the whole text description (a sequence of words).

We can first calculate the similarity matrix between all possible pairs of word and image region by Eq.~(\ref{eq:similarity}),
\begin{equation}\label{eq:similarity}
s = \dot{e}^T\,v,
\end{equation}
where $s \in \mathbb{R}^{T\times 289}$ and $s_{i,j}$ means the similarity between the $i^{th}$ word and the $j^{th}$ image region.

Generally, a sub-region of the image is described by none or several words of the text description, and it is not likely to be described by the whole sentence.
Therefore, we normalize the similarity matrix by Eq.~(\ref{eq:prob}),
 \begin{equation}\label{eq:prob}
  \overline{s}_{i,j} = \frac{\exp(s_{i,j})}{\sum_{k=0}^{T-1}{\exp(s_{k,j})}}
 \end{equation}
% When an image region (\eg, \textit{red pointed beak}) is described by multiple words of the text description, the normalized similarities between this region and its relevant words are lager than $1/T$ while the normalized similarities between this region and its non-relevant words are smaller than $1/T$. When an image region, such as \textit{sky} in CUB dataset, is not related to any word of the description, the normalized similarities between this region and all words are around $1/T$.

Second, we build an attention model to compute a context vector for each word (query).
The context vector $c_{i}$ is a dynamic representation of image regions related to the $i^{th}$ word of the text description.
It is computed as the weighted sum over all visual feature vectors,
 \begin{equation}\label{eq:attention}
  c_i = \sum_{j=0}^{288}\alpha_{j}v_{j},
 \end{equation}
where we define the weight $\alpha_{j}$ via Eq.~(\ref{eq:weight}),
 \begin{equation}\label{eq:weight}
  \alpha_{j} = \frac{\exp(\gamma_{1} \overline{s}_{i,j})}{\sum_{k=0}^{288}{\exp(\gamma_{1} \overline{s}_{i,k}})}
 \end{equation}
Here, $\gamma_{1}$ is a factor that decides how much more attention is paid to features of its relevant regions when computing the context vector for a word.
% Higher values for $\gamma_{1} \overline{s}_{i,j}$ result in higher weights and vice versa.

Finally, we define the relevance between the $i^{th}$ word and the image using the cosine similarity between $c_i$ and $\dot{e}_i$, \ie, $R(c_i, \dot{e}_i) = (c_i^T \dot{e}_i) / (||c_i|| ||\dot{e}_i||)$.
The relevance between the entire image (Q) and the whole text description (U) is computed by Eq.~(\ref{eq:relevance}),
 \begin{equation}\label{eq:relevance}
  R(Q, U) = \log \Big(\sum_{i=1}^{T-1} \exp(\gamma_{2} R(c_i, \dot{e}_i)) \Big)^{\frac{1}{\gamma_{2}}},
 \end{equation}
where $\gamma_{2}$ is a factor that determines how much to magnify the importance of the most relevant word-image pair. When $\gamma_{2}\to\infty$, $R(Q, U)$ approximates to $\max_{i=1}^{T-1} R(c_i, \dot{e}_i)$.

For a text-image pair, we can compute the posterior probability
of the text description ($U$) being matching with the image ($Q$) via,
 \begin{equation}\label{eq:relevant_prob}
  P(U|Q) = \frac{\exp(\gamma_{3} R(Q, U))}{\sum_{U^{\prime}\in\mathbb{U}}{\exp(\gamma_{3} R(Q, U^{\prime}))}},
 \end{equation}
where $\gamma_{3}$ is a smoothing factor determined by experiments. $\mathbb{U}$ denotes a minibatch of $M$ text descriptions, in which only one description $U^+$ matches the image $Q$.
Thus, for each image, there are $M-1$ mismatching text descriptions.
The objective function is to learn the model
parameters $\Lambda$ by minimizing the negative log posterior probability
that the images are matched with their corresponding text descriptions (ground truth),
\begin{equation}\label{eq:L1}
\mathcal{L}^{w}_{1}(\Lambda) = -\log \prod_{Q\in\mathbb{Q}} P(U^+|Q),
\end{equation}
where `w' stands for ``word''.

Symmetrically, we can compute,
\begin{equation}\label{eq:L2}
\mathcal{L}^{w}_{2}(\Lambda) = -\log \prod_{U\in\mathbb{U}} P(Q^+|U),
\end{equation}
where $P(Q|U) = \frac{\exp(\gamma_{3} R(Q, U))}{\sum_{Q^{\prime}\in\mathbb{Q}}{\exp(\gamma_{3} R(Q^{\prime}, U))}}$.

If we redefine Eq.~(\ref{eq:relevance}) by $R(Q, U) = \big(\overline{v}^T \hat{e}\big) / \big(||\overline{v}|| ||\hat{e}||\big)$ and substitute it to Eq.~(\ref{eq:relevant_prob}), Eq.~(\ref{eq:L1}), Eq.~(\ref{eq:L2}), we can obtain loss functions $\mathcal{L}^{s}_{1}$ and $\mathcal{L}^{s}_{2}$ (where `s' stands for ``sentence'') using the sentence embedding $\hat{e}$ and the global visual vector $\overline{v}$.

The fine-grained conditional loss is defined via Eq.~(\ref{eq:L_DAMSM}),
 \begin{equation}\label{eq:L_DAMSM}
  % \begin{aligned}
   \mathcal{L}_{DAMSM} = \mathcal{L}^{w}_{1} + \mathcal{L}^{w}_{2} + \mathcal{L}^{s}_{1} + \mathcal{L}^{s}_{2}
  % \end{aligned}
 \end{equation}
The DAMSM is pre-trained by minimizing $\mathcal{L}_{DAMSM}$ using real image-text pairs.
Since the size of images for pre-training DAMSM is not limited by the size of images that can be generated, real images of size 299$\times$299 are utilized.
Furthermore, the pre-trained DAMSM can provide visually-discriminative word features and a stable fine-grained conditional loss for the attention generative network.
}

\textbf{The R-precision score.}{
The DAMSM model is also used to compute the R-precision score. If there are $R$ relevant documents for a query, we examine the top $R$ ranked retrieval results of a system, and find that $r$ are relevant, and then by definition, the R-precision (and also the precision and recall) is ${r}/{R}$.
More specifically, we use generated images to query their corresponding text descriptions.
First, the image encoder and Bi-LSTM text encoder learned in our pre-trained DAMSM are utilized to extract features of the generated images and the given text descriptions.
Then, we compute cosine similarities between the image features and the text features.
Finally, we rank candidates text descriptions for each image in descending similarity and find the top $r$ relevant descriptions for computing the R-precision.
}

\subsection{Network architectures for semantic layout generation}
\label{sec:layout}

\textbf{Box generator.}
We design our box generator by improving the one in \cite{hong2018inferring} to be attentive. We denote the bounding box of the $t$-th object as $B_{t} = (b^{x}_{t}, b^{y}_{t}, b^{w}_{t}, b^{h}_{t}, \bm{l}_{t})$. Then, we formulate the joint probability of sampling $B_{t}$ from the box generator as
\begin{equation}\label{eqn:boxgen}
    p(b^{x}_{t}, b^{y}_{t}, b^{w}_{t}, b^{h}_{t}, \bm{l}_{t}) = p(\bm{l}_{t})p(b^{x}_{t}, b^{y}_{t}, b^{w}_{t}, b^{h}_{t}|\bm{l}_{t}).
\end{equation}
We implement $p(\bm{l}_{t})$ as a categorical distribution, and implement $p(b^{x}_{t}, b^{y}_{t}, b^{w}_{t}, b^{h}_{t}|\bm{l}_{t})$ as a mixture of quadravariate Gaussians. As described in \cite{hong2018inferring}, in order to reduce the parameter space, we decompose the box coordinate probability as $p(b^{x}_{t}, b^{y}_{t}, b^{w}_{t}, b^{h}_{t}|\bm{l}_{t}) = p(b^{x}_{t}, b^{y}_{t}|\bm{l}_{t})p(b^{w}_{t}, b^{h}_{t}|b^{x}_{t}, b^{y}_{t},\bm{l}_{t})$, and approximate it with two bivariate Gasussian mixtures by
%\begin{align}
%  &\begin{aligned}
%    &\mathllap{p(b^{x}_{t}, b^{y}_{t}|\bm{l}_{t})} = \sum_{k=1}^{K} \pi_{t,k}^{xy}\mathcal{N}(b^{x}_{t}, b^{y}_{t}; \bm{\mu}^{xy}_{t,k}, \Sigma^{xy}_{t,k}),
%  \end{aligned}\\
%  &\begin{aligned}
%    &\mathllap{p(b^{w}_{t}, b^{h}_{t}|b^{x}_{t}, b^{y}_{t}, \bm{l}_{t})} = \sum_{k=1}^{K} \pi_{t,k}^{wh}\mathcal{N}(b^{w}_{t}, b^{h}_{t}; \bm{\mu}^{wh}_{t,k}, \Sigma^{wh}_{t,k}).
%  \end{aligned}
%\end{align}
\begin{equation}
\begin{split}
    p(b^{x}_{t}, b^{y}_{t}|\bm{l}_{t}) &= \sum_{k=1}^{K} \pi_{t,k}^{xy}\mathcal{N}(b^{x}_{t}, b^{y}_{t}; \bm{\mu}^{xy}_{t,k}, \Sigma^{xy}_{t,k}),\\
    p(b^{w}_{t}, b^{h}_{t}|b^{x}_{t}, b^{y}_{t}, \bm{l}_{t}) &= \sum_{k=1}^{K} \pi_{t,k}^{wh}\mathcal{N}(b^{w}_{t}, b^{h}_{t}; \bm{\mu}^{wh}_{t,k}, \Sigma^{wh}_{t,k}).
\end{split}
\end{equation}
In practice, as in \cite{hong2018inferring}, we implement the box generator within a encoder-decoder framework. The encoder is the Bi-LSTM text encoder as mentioned in $\S$~\ref{sec:DAMSM}. The Gaussian Mixture Model (GMM) parameters for Eq. \eqref{eqn:boxgen} are obtained from the decoder LSTM outputs. Given text encoder's final hidden state $h^{\textrm{Enc}}_{T_s} \in \mathbb{R}^{D}$ and output $H^{\textrm{Enc}} \in \mathbb{R}^{T_s \times D}$, we initialize the decoder's initial hidden state $h_{0}$ with $h^{\textrm{Enc}}_{T_s}$. As for $H^{\textrm{Enc}}$, we use it to compute the contextual input $z_{t}$ for the decoder:
\begin{align}
  &\begin{aligned}
    \mathllap{z_{t}} &= \sum_{i=1}^{T_s} \alpha_{i} h^{\textrm{Enc}}_{i}, ~\textrm{with}~ \alpha_{i}=W_{v} \cdot (W_{\alpha}[h_{t-1}, h^{\textrm{Enc}}_{i}]),
  \end{aligned}
\end{align}
where $W_{v}$ is a learnable parameter, $W_{\alpha}$ is the parameter of a linear transformation, and $\cdot$ and $[\cdot, \cdot]$ represent the dot product and concatenation operation, respectively.

Then, the calculation of GMM parameters are shown as follows:
\begin{align}
  &\begin{aligned}
    \mathllap{[h_{t}, c_{t}]} &= \textrm{LSTM}([B_{t-1}, z_{t}]; [h_{t-1}, c_{t-1}]),
  \end{aligned}\\
  &\begin{aligned}
    \mathllap{\bm{l}_{t}} &= W^{l}h_{t}+\mathbf{b}^{l},
  \end{aligned}\\
  &\begin{aligned}
    \mathllap{\bm{\theta}^{xy}_{t}} &= W^{xy}[h_{t}, \bm{l}_{t}]+\mathbf{b}^{xy},
  \end{aligned}\\
  &\begin{aligned}
    \mathllap{\bm{\theta}^{wh}_{t}} &= W^{wh}[h_{t}, \bm{l}_{t}, b_{x}, b_{y}]+\mathbf{b}^{wh},
  \end{aligned}
\end{align}
where $\bm{\theta}^{\cdot}_{t} = [\bm{\pi}^{\cdot}_{t,1:K}, \bm{\mu}^{\cdot}_{t,1:K}, \bm{\Sigma}^{\cdot}_{t,1:K}]$ are the parameters for GMM concatenated to a vector.
We use the same Adam optimizer and training hyperparameters (\ie, learning rate $0.001$, $\beta_{1}=0.9$, $\beta_{2}=0.999$) as in \cite{hong2018inferring}.

\textbf{Shape generator.} We implement the shape generator in \cite{hong2018inferring} with almost the same architecture except the upsample block. In \cite{hong2018inferring}, the upsample block is designed as [convtranspose $4 \times 4$ (pad $1$, stride $2$) - Instance Normalization - ReLU]. We discovered that the usage of convtranspose would lead to unstable training which is reflected by the frequent severe grid artifacts. To this end, we replace this upsample block with that in our image generator (see Table~\ref{tab:basic_blocks}) by switching the batch normalization to the instance one.

\subsection{Network architectures for image generation}

We present the network architecture for image generators in Table~\ref{tab:gen_struct} and the network architectures for discriminators in Table~\ref{tab:patd_struct}, Table~\ref{tab:shpd_struct} and Table~\ref{tab:objd_struct}. They are built with basic blocks defined in Table~\ref{tab:basic_blocks}. We set the hyperparameters of the network structures as: $N_{g}=48$, $N_{d}=96$, $N_{c}=80$, $N_{e}=256$, $N_{l}=50$, $m_{0}=7$, $m_{1}=3$, and $m_{2}=3$.

We employ an Adam optimizer for the generators with learning rate $0.0002$, $\beta_{1} = 0.5$ and $\beta_{2} = 0.999$. For each discriminator, we also employ an Adam optimizer with the same hyperparameters.

We design the object-wise discriminators for small objects and large objects, respectively. We specify that if the maximum of width or height of an object is greater than one-third of the image size, then this object is large; otherwise, it is small.

\subsection{Network architectures for spectral normalized projection discriminators}

\begin{figure*}[h]
\footnotesize
        \centering
        \vspace{-0.2cm}
        \begin{subfigure}[b]{0.45\textwidth}
                \includegraphics[width=\textwidth]{discriminator}
                \caption{Object-wise discriminator.}
        \end{subfigure}
        ~
        \begin{subfigure}[b]{0.45\textwidth}
                \includegraphics[width=\textwidth]{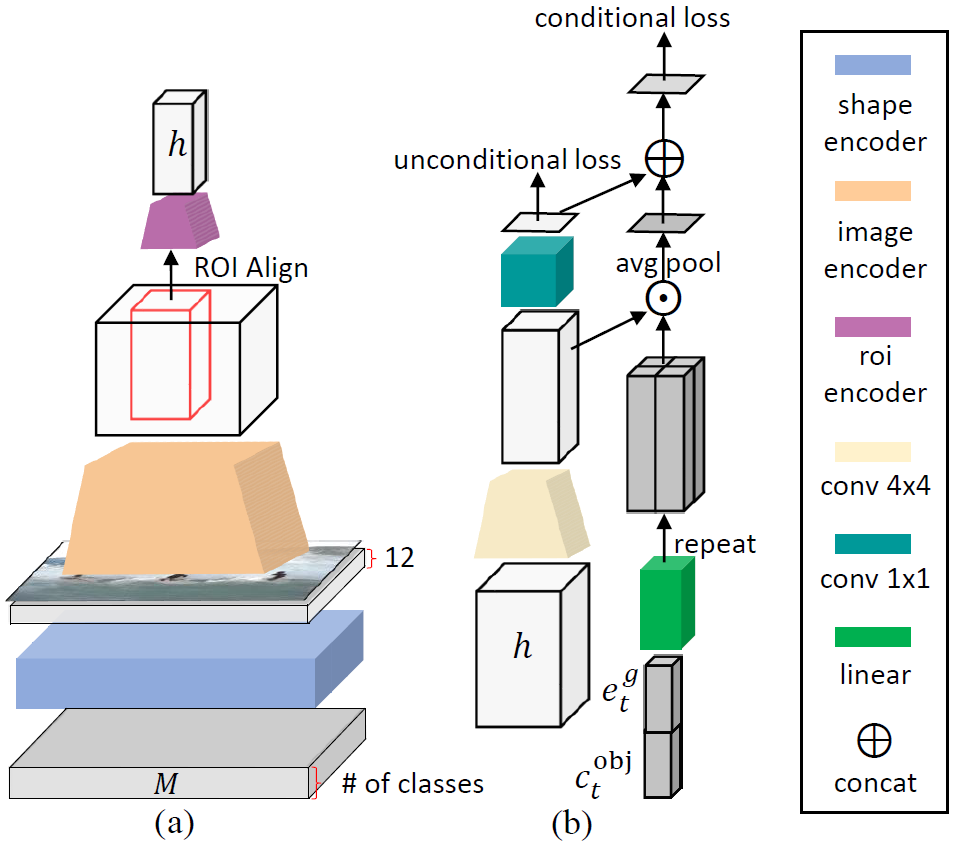}
                \caption{Object-wise spectral normalized projection discriminator.}
        \end{subfigure}
        \vspace{-0.2cm}
        \caption{The comparison between the object-wise discriminator and its spectral normalized projection version. (a) extracts the region feature based on the Fast R-CNN model. (b) determines whether the $t$-th object is realistic (consistent with its label $e^{\textrm{g}}_{t}$ and text context information $c^{\textrm{obj}}_{t}$) or not.}
        \vspace{-0.2cm}
    \label{fig:comp_w_pjd}
\end{figure*}

We combine our discriminators above with the spectral normalized projection discriminator in \cite{miyato2018spectral,miyato2018cgans}. The difference between the object-wise discriminator and the object-wise spectral normalized projection discriminator is illustrated in Figure~\ref{fig:comp_w_pjd}. We present detailed network architectures of the spectral normalized projection discriminators in Table~\ref{tab:pjpatd_struct}, Table~\ref{tab:pjshpd_struct} and Table~\ref{tab:pjobjd_struct}, with  basic blocks defined in Table~\ref{tab:basic_blocks}.

\begin{table*}[h]
\footnotesize
\centering
\caption{The architecture of Bi-LSTM text encoder.}
\begin{tabular}[t]{|p{1.6cm}|p{12.8cm}|}\hline
{Layer Name} &{\makebox[12.8cm]{Hyperparameters}}\\ \hline\hline
{Embedding} &$\textrm{num embeddings}=\textrm{vocab size}$, $\textrm{embedding dim}=300$ \\ \hline
{Dropout} &$\textrm{prob}=0.5$ \\ \hline
{LSTM} &$\textrm{input size}=300$, $\textrm{hidden size}~(\frac{D}{2}) =128$, $\textrm{num layers}=1$, $\textrm{dropout prob}=0.5$, $\textrm{bidirectional}=\textrm{True}$ \\ \hline

\end{tabular}
\label{tab:bilstm}
\end{table*}

\begin{table*}[h]
\footnotesize
\centering
\caption{The basic blocks for architecture design. (``-" connects two consecutive layers; ``+" means element-wise addition between two layers.)}
\begin{tabular}[t]{|p{3.3cm}|p{13.3cm}|}\hline
{Name} &{\makebox[13.3cm]{Operations / Layers}}\\ \hline\hline
{Interpolating ($k$)} &Nearest neighbor upsampling layer (up-scaling the spatial size by $k$) \\ \hline
\multirow{ 2}{*}{Upsampling ($k$)} &Interpolating ($2$) - convolution $3 \times 3$ (stride $1$, padding $1$, decreasing $\sharp$channels to $k$) - \\
&Batch Normalization (BN) - Gated Linear Unit (GLU). \\ \hline
\multirow{ 2}{*}{Downsampling ($k$)} &In $G$s: convolution $3 \times 3$ (stride $2$, increasing $\sharp$channels to $k$) - BN - LeakyReLU. \\
&In $D$s, the convolutional kernel size is $4$. In the first block of $D$s, BN is not applied. \\ \hline
\multirow{2}{*}{Downsampling w/ SN ($k$)} &Convolution $4 \times 4$ (spectral normalized, stride $2$, increasing $\sharp$channels to $k$) - BN - LeakyReLU.\\
& In the first block of $D$s, BN is not applied.\\\hline
Concat &Concatenate input tensors along the channel dimension.  \\ \hline
\multirow{ 2}{*}{Residual} &Input $+$ [Reflection Pad (RPad) 1 - convolution $3 \times 3$ (stride $1$, doubling $\sharp$channels) -\\
&Instance Normalization (IN) - GLU - RPad 1 - convolution $3 \times 3$ (stride $1$, keeping $\sharp$channels) - IN]. \\ \hline
FC &At the beginning of $G$s: fully connected layer - BN - GLU - reshape to 3D tensor.  \\\hline
FC w/ SN ($k$) &Fully connected layer (spectral normalized, changing $\sharp$channels to $k$).  \\\hline
{Outlogits} &Convolution $4 \times 4$ (stride $2$, decreasing $\sharp$channels to $1$) - sigmoid.\\\hline
{Repeat ($k \times k$)} &Copy a vector $k \times k$ times.\\\hline
Fmap Sum &Summing the two input feature maps element-wisely. \\\hline
Fmap Mul &Multiplying the two input feature maps element-wisely. \\\hline
Avg Pool ($k$) &Average pooling along the $k$-th dimension. \\\hline
\multirow{ 2}{*}{Conv $3 \times 3~(k)$} &In $G$s: convolution $3 \times 3$ (stride $1$, padding $1$, changing $\sharp$channels to $k$) - Tanh.\\
&In $D$s, convolution $3 \times 3$ (stride $1$, padding $1$, changing $\sharp$channels to $k$) - BN - LeakyReLU.\\\hline
{Conv $4 \times 4$ w/ SN} &Convolution $4 \times 4$ (spectral normalized, stride $2$, keeping $\sharp$channels).\\\hline
{Conv $1 \times 1$ w/ SN} &Convolution $1 \times 1$ (spectral normalized, stride $1$, decreasing $\sharp$channels to $1$).\\\hline
\multirow{ 2}{*}{$F^{\textrm{ca}}$} &Conditioning augmentation that converts the sentence embedding $\hat{e}$ to the conditioning vector $\overline{e}$:\\
&fully connected layer - ReLU.  \\\hline
$F^{\textrm{pat-attn}}$ &Grid attention module. Refer to the paper for more details. \\\hline
$F^{\textrm{obj-attn}}$ &Object-driven attention module. Refer to the paper for more details. \\\hline
$F^{\textrm{lab-distr}}$ &Label distribution module. Refer to the paper for more details. \\\hline
Shape Encoder ($k$) &RPad 1 - convolution $3 \times 3$ (stride $1$, decreasing $\sharp$channels to $k$) - IN - LeakyReLU.\\\hline
Shape Encoder w/ SN ($k$) &RPad 1 - convolution $3 \times 3$ (spectral normalized, stride $1$, decreasing $\sharp$channels to $k$) - IN - LeakyReLU.\\\hline
ROI Encoder &Convolution $4 \times 4$ (stride $1$, padding $1$, decreasing $\sharp$channels to $N_{d}*4$) - LeakyReLU.  \\\hline
ROI Encoder w/ SN &Convolution $4 \times 4$ (spectral normalized, stride $1$, padding $1$, decreasing $\sharp$channels to $N_{d}*4$) - LeakyReLU.  \\\hline
ROI Align ($k$) &Pooling $k \times k$ feature maps for ROI.  \\\hline

\end{tabular}
\label{tab:basic_blocks}
\end{table*}

\begin{table*}[h]
\footnotesize
\centering
\caption{The structure for generators of Obj-GAN.}
\begin{tabular}[t]{|c|p{2.8cm}|c|c|}\hline
{Stage} &{Name} &{\makebox[4cm]{Input Tensors}} &{\makebox[4cm]{Output Tensors}}\\ \hline\hline
\multirow{9}{*}{$G_{0}$} &FC &100-dimensional $z$, and $F^{\textrm{ca}}$ & $8 \times 8 \times 4N_{g}$ \\\cline{2-4}
&Upsampling ($2N_{g}$) & $8 \times 8 \times 4N_{g}$ & $16 \times 16 \times 2N_{g}$ \\\cline{2-4}
&Upsampling ($N_{g}$) & $16 \times 16 \times 2N_{g}$ & $c$ ($32 \times 32\times N_{g}$) \\\cline{2-4}
&Shape Encoder ($\frac{1}{2}N_{g}$) & $M^{0}$ ($64 \times 64 \times N_{c}$) & $64 \times 64\times \frac{1}{2}N_{g}$ \\\cline{2-4}
&Downsampling ($N_{g}$) & $64 \times 64 \times \frac{1}{2}N_{g}$ & $u_{0}$ ($32 \times 32 \times N_{g}$) \\\cline{2-4}
&Concat & $c, u_{0}, F^{\textrm{obj-attn}}, F^{\textrm{lab-distr}}$ & $32 \times 32 \times (3N_{g}+N_{l})$ \\\cline{2-4}
&$m_{0}$ Residual & $32 \times 32 \times (3N_{g}+N_{l})$ & $32 \times 32 \times (3N_{g}+N_{l})$ \\\cline{2-4}
&Upsampling ($N_{g}$) & $32 \times 32 \times (3N_{g}+N_{l})$ & $h_{0}$ ($64 \times 64\times N_{g}$) \\\cline{2-4}
&Conv $3 \times 3~(3)$ & $h_{0}$ & $x_{0}$ ($64 \times 64\times 3$) \\\hline\hline

\multirow{7}{*}{$G_{1}$} &Shape Encoder ($\frac{1}{2}N_{g}$) & $M^{1}$ ($128 \times 128 \times N_{c}$) & $128 \times 128\times \frac{1}{2}N_{g}$ \\\cline{2-4}
&Downsampling ($N_{g}$) & $128 \times 128 \times \frac{1}{2}N_{g}$ & $u_{1}$ ($64 \times 64 \times N_{g}$) \\\cline{2-4}
&Fmap Sum & $h_{0}, u_{1}$ & $h_{0}$ ($64 \times 64 \times N_{g}$) \\\cline{2-4}
&Concat & $F^{\textrm{pat-attn}}, h_{0}, F^{\textrm{obj-attn}}, F^{\textrm{lab-distr}}$ & $64 \times 64 \times (3N_{g}+N_{l})$ \\\cline{2-4}
&$m_{1}$ Residual & $64 \times 64 \times (3N_{g}+N_{l})$ & $64 \times 64 \times (3N_{g}+N_{l})$ \\\cline{2-4}
&Upsampling ($N_{g}$) & $64 \times 64 \times (3N_{g}+N_{l})$ & $h_{1}$ ($128 \times 128\times N_{g}$) \\\cline{2-4}
&Conv $3 \times 3~(3)$ & $h_{1}$ & $x_{1}$ ($128 \times 128\times 3$) \\\hline\hline

\multirow{7}{*}{$G_{2}$} &Shape Encoder ($\frac{1}{2}N_{g}$) & $M^{2}$ ($256 \times 256 \times N_{c}$) & $256 \times 256\times \frac{1}{2}N_{g}$ \\\cline{2-4}
&Downsampling ($N_{g}$) & $256 \times 256 \times \frac{1}{2}N_{g}$ & $u_{2}$ ($128 \times 128 \times N_{g}$) \\\cline{2-4}
&Fmap Sum & $h_{1}, u_{2}$ & $h_{1}$ ($128 \times 128 \times N_{g}$) \\\cline{2-4}
&Concat & $F^{\textrm{pat-attn}}, h_{1}, F^{\textrm{obj-attn}}, F^{\textrm{lab-distr}}$ & $128 \times 128 \times (3N_{g}+N_{l})$ \\\cline{2-4}
&$m_{2}$ Residual & $128 \times 128 \times (3N_{g}+N_{l})$ & $128 \times 128 \times (3N_{g}+N_{l})$ \\\cline{2-4}
&Upsampling ($N_{g}$) & $128 \times 128 \times (3N_{g}+N_{l})$ & $h_{2}$ ($256 \times 256\times N_{g}$) \\\cline{2-4}
&Conv $3 \times 3~(3)$ & $h_{2}$ & $x_{2}$ ($256 \times 256\times 3$) \\\hline
\end{tabular}
\label{tab:gen_struct}
\end{table*}

\begin{table*}[h]
\footnotesize
\centering
\caption{The structure for patch-wise discriminators of Obj-GAN. $\overline{e}$ is output by $F^{\textrm{ca}}$}
\begin{tabular}[t]{|c|p{3.8cm}|c|c|}\hline
{Stage} &{Name} &{\makebox[4cm]{Input Tensors}} &{\makebox[4cm]{Output Tensors}}\\ \hline\hline
\multirow{7}{*}{$D_{0}$} &Downsampling ($N_{d}$) & $x_{0}~(64 \times 64 \times 3)$ & $32 \times 32 \times N_{d}$ \\\cline{2-4}
&Downsampling ($2N_{d}$) & $32 \times 32 \times N_{d}$ & $16 \times 16 \times 2N_{d}$ \\\cline{2-4}
&Downsampling ($4N_{d}$) & $16 \times 16 \times 2N_{d}$ & $8 \times 8 \times 4N_{d}$ \\\cline{2-4}
&Downsampling ($8N_{d}$) & $8 \times 8 \times 4N_{d}$ & $h_{0}$ ($4 \times 4 \times 8N_{d})$ \\\cline{2-4}
&Repeat ($4 \times 4$) &$\overline{e}~(N_{e})$ & $4 \times 4 \times N_{e}$ \\\cline{2-4}
&Concat - Conv $3 \times 3~(8N_{d})$ & $h_{0}, 4 \times 4 \times N_{e}$  & $he_{0}$ ($4 \times 4 \times 8N_{d}$)\\\cline{2-4}
&Outlogits (unconditional loss) & $h_{0}$  & $1$\\\cline{2-4}
&Outlogits (conditional loss) & $he_{0}$  & $1$\\\hline

\multirow{7}{*}{$D_{1}$} &Downsampling ($N_{d}$) & $x_{1}~(128 \times 128 \times 3)$ & $64 \times 64 \times N_{d}$ \\\cline{2-4}
&Downsampling ($2N_{d}$) & $64 \times 64 \times N_{d}$ & $32 \times 32 \times 2N_{d}$ \\\cline{2-4}
&Downsampling ($4N_{d}$) & $32 \times 32 \times 2N_{d}$ & $16 \times 16 \times 4N_{d}$ \\\cline{2-4}
&Downsampling ($8N_{d}$) & $16 \times 16 \times 4N_{d}$ & $h_{1}$ ($8 \times 8 \times 8N_{d})$ \\\cline{2-4}
&Repeat ($8 \times 8$) &$\overline{e}~(N_{e})$ & $8 \times 8 \times N_{e}$ \\\cline{2-4}
&Concat - Conv $3 \times 3~(8N_{d})$ & $h_{1}, 8 \times 8 \times N_{e}$  & $he_{1}$ ($8 \times 8 \times 8N_{d}$)\\\cline{2-4}
&Outlogits (unconditional loss) & $h_{1}$  & $3 \times 3$\\\cline{2-4}
&Outlogits (conditional loss) & $he_{1}$  & $3 \times 3$\\\hline

\multirow{7}{*}{$D_{2}$} &Downsampling ($N_{d}$) & $x_{2}~(256 \times 256 \times 3)$ & $128 \times 128 \times N_{d}$ \\\cline{2-4}
&Downsampling ($2N_{d}$) & $128 \times 128 \times N_{d}$ & $64 \times 64 \times 2N_{d}$ \\\cline{2-4}
&Downsampling ($4N_{d}$) & $64 \times 64 \times 2N_{d}$ & $32 \times 32 \times 4N_{d}$ \\\cline{2-4}
&Downsampling ($8N_{d}$) & $32 \times 32 \times 4N_{d}$ & $h_{2}$ ($16 \times 16 \times 8N_{d})$ \\\cline{2-4}
&Repeat ($16 \times 16$) &$\overline{e}~(N_{e})$ & $16 \times 16 \times N_{e}$ \\\cline{2-4}
&Concat - Conv $3 \times 3~(8N_{d})$ & $h_{2}, 16 \times 16 \times N_{e}$  & $he_{2}$ ($16 \times 16 \times 8N_{d}$)\\\cline{2-4}
&Outlogits (unconditional loss) & $h_{2}$  & $7 \times 7$\\\cline{2-4}
&Outlogits (conditional loss) & $he_{2}$  & $7 \times 7$\\\hline

\end{tabular}
\label{tab:patd_struct}
\end{table*}

\begin{table*}[h]
\footnotesize
\centering
\caption{The structure for shape discriminators of Obj-GAN.}
\begin{tabular}[t]{|c|p{3.8cm}|c|c|}\hline
{Stage} &{Name} &{\makebox[4cm]{Input Tensors}} &{\makebox[4cm]{Output Tensors}}\\ \hline\hline
\multirow{7}{*}{$D_{0}$} &Shape Encoder ($\frac{1}{8}N_{d}$) & $M^{0}$ ($64 \times 64 \times N_{c}$) &$64 \times 64\times \frac{1}{8}N_{d}$ \\\cline{2-4}
&Concat & $x_{0}~(64 \times 64 \times 3), 64 \times 64\times \frac{1}{8}N_{d}$ & $64 \times 64 \times (3+\frac{1}{8}N_{d})$ \\\cline{2-4}
&Downsampling ($N_{d}$) & $64 \times 64 \times (3+\frac{1}{8}N_{d})$ & $32 \times 32 \times N_{d}$ \\\cline{2-4}
&Downsampling ($2N_{d}$) & $32 \times 32 \times N_{d}$ & $16 \times 16 \times 2N_{d}$ \\\cline{2-4}
&Downsampling ($4N_{d}$) & $16 \times 16 \times 2N_{d}$ & $8 \times 8 \times 4N_{d}$ \\\cline{2-4}
&Downsampling ($8N_{d}$) & $8 \times 8 \times 4N_{d}$ & $h_{0}$ ($4 \times 4 \times 8N_{d})$ \\\cline{2-4}
&Outlogits (unconditional loss) & $h_{0}$  & $1$\\\hline

\multirow{7}{*}{$D_{1}$} &Shape Encoder ($\frac{1}{8}N_{d}$) & $M^{1}$ ($128 \times 128 \times N_{c}$) &$128 \times 128 \times \frac{1}{8}N_{d}$ \\\cline{2-4}
&Concat & $x_{1}~(128 \times 128 \times 3), 128 \times 128 \times \frac{1}{8}N_{d}$ & $128 \times 128 \times (3+\frac{1}{8}N_{d})$ \\\cline{2-4}
&Downsampling ($N_{d}$) & $128 \times 128 \times (3+\frac{1}{8}N_{d})$ & $64 \times 64 \times N_{d}$ \\\cline{2-4}
&Downsampling ($2N_{d}$) & $64 \times 64 \times N_{d}$ & $32 \times 32 \times 2N_{d}$ \\\cline{2-4}
&Downsampling ($4N_{d}$) & $32 \times 32 \times 2N_{d}$ & $16 \times 16 \times 4N_{d}$ \\\cline{2-4}
&Downsampling ($8N_{d}$) & $16 \times 16 \times 4N_{d}$ & $h_{1}$ ($8 \times 8 \times 8N_{d})$ \\\cline{2-4}
&Outlogits (unconditional loss) & $h_{1}$  & $3 \times 3$\\\hline

\multirow{7}{*}{$D_{2}$} &Shape Encoder ($\frac{1}{8}N_{d}$) & $M^{2}$ ($256 \times 256 \times N_{c}$) &$256 \times 256 \times \frac{1}{8}N_{d}$ \\\cline{2-4}
&Concat & $x_{2}~(256 \times 256 \times 3), 256 \times 256 \times \frac{1}{8}N_{d}$ & $256 \times 256 \times (3+\frac{1}{8}N_{d})$ \\\cline{2-4}
&Downsampling ($N_{d}$) & $256 \times 256 \times (3+\frac{1}{8}N_{d})$ & $128 \times 128 \times N_{d}$ \\\cline{2-4}
&Downsampling ($2N_{d}$) & $128 \times 128 \times N_{d}$ & $64 \times 64 \times 2N_{d}$ \\\cline{2-4}
&Downsampling ($4N_{d}$) & $64 \times 64 \times 2N_{d}$ & $32 \times 32 \times 4N_{d}$ \\\cline{2-4}
&Downsampling ($8N_{d}$) & $32 \times 32 \times 4N_{d}$ & $h_{2}$ ($16 \times 16 \times 8N_{d})$ \\\cline{2-4}
&Outlogits (unconditional loss) & $h_{2}$  & $7 \times 7$\\\hline

\end{tabular}
\label{tab:shpd_struct}
\end{table*}

\begin{table*}[h]
\footnotesize
\centering
\caption{The structure for object-wise discriminators of Obj-GAN. $c^{\textrm{obj}}$ represents the intermediate context vectors of $F^{\textrm{obj-attn}}$, and $e^{\textrm{g}}$ represents the embedding vectors the class labels.}
\begin{tabular}[t]{|c|p{3.8cm}|c|c|}\hline
{Stage} &{Name} &{\makebox[4cm]{Input Tensors}} &{\makebox[4cm]{Output Tensors}}\\ \hline\hline
\multirow{12}{*}{small} &Interpolating ($2$) & $M^{2}$ ($256 \times 256 \times N_{c}$) &$512 \times 512 \times N_{c}$ \\\cline{2-4}
&Interpolating ($2$) & $x^{2}$ ($256 \times 256 \times 3$) &$512 \times 512 \times 3$ \\\cline{2-4}
&Shape Encoder ($\frac{1}{8}N_{d}$) & $512 \times 512 \times N_{c}$ &$512 \times 512 \times \frac{1}{8}N_{d}$ \\\cline{2-4}
&Concat & $512 \times 512 \times 3, 512 \times 512 \times \frac{1}{8}N_{d}$ & $512 \times 512 \times (3+\frac{1}{8}N_{d})$ \\\cline{2-4}
&Downsampling ($N_{d}$) & $512 \times 512 \times (3+\frac{1}{8}N_{d})$ & $256 \times 256 \times N_{d}$ \\\cline{2-4}
&Downsampling ($2N_{d}$) & $256 \times 256 \times N_{d}$ & $128 \times 128 \times 2N_{d}$ \\\cline{2-4}
&Downsampling ($4N_{d}$) & $128 \times 128 \times 2N_{d}$ & $64 \times 64 \times 4N_{d}$ \\\cline{2-4}
&ROI Align ($5$) & $64 \times 64 \times 4N_{d}$ & $N_{\textrm{small}} \times 5 \times 5 \times 4N_{d}$ \\\cline{2-4}
&ROI Encoder ($5$) & $N_{\textrm{small}} \times 5 \times 5 \times 4N_{d}$ & $h~(N_{\textrm{small}} \times 4 \times 4 \times 4N_{d})$ \\\cline{2-4}
&Repeat ($4 \times 4$) &$c^{\textrm{obj}}~(N_{\textrm{small}} \times N_{g})$ & $N_{\textrm{small}} \times 4 \times 4 \times N_{g}$ \\\cline{2-4}
&Repeat ($4 \times 4$) &$e^{\textrm{g}}~(N_{\textrm{small}} \times N_{l})$ & $N_{\textrm{small}} \times 4 \times 4 \times N_{l}$ \\\cline{2-4}
&Concat - Conv $3 \times 3~(4N_{d})$ & $h, N_{\textrm{small}} \times 4 \times 4 \times N_{g}, N_{\textrm{small}} \times 4 \times 4 \times N_{l}$  & $hc$ ($N_{\textrm{small}} \times 4 \times 4 \times 4N_{d}$)\\\cline{2-4}
&Outlogits (unconditional loss) & $h$  & $N_{\textrm{small}}$\\\cline{2-4}
&Outlogits (conditional loss) & $hc$  & $N_{\textrm{small}}$\\\hline\hline

\multirow{13}{*}{large} &Interpolating ($2$) & $M^{2}$ ($256 \times 256 \times N_{c}$) &$512 \times 512 \times N_{c}$ \\\cline{2-4}
&Interpolating ($2$) & $x^{2}$ ($256 \times 256 \times 3$) &$512 \times 512 \times 3$ \\\cline{2-4}
&Shape Encoder ($\frac{1}{8}N_{d}$) & $512 \times 512 \times N_{c}$ &$512 \times 512 \times \frac{1}{8}N_{d}$ \\\cline{2-4}
&Concat & $512 \times 512 \times 3, 512 \times 512 \times \frac{1}{8}N_{d}$ & $512 \times 512 \times (3+\frac{1}{8}N_{d})$ \\\cline{2-4}
&Downsampling ($N_{d}$) & $512 \times 512 \times (3+\frac{1}{8}N_{d})$ & $256 \times 256 \times N_{d}$ \\\cline{2-4}
&Downsampling ($2N_{d}$) & $256 \times 256 \times N_{d}$ & $128 \times 128 \times 2N_{d}$ \\\cline{2-4}
&Downsampling ($4N_{d}$) & $128 \times 128 \times 2N_{d}$ & $64 \times 64 \times 4N_{d}$ \\\cline{2-4}
&Downsampling ($8N_{d}$) & $64 \times 64 \times 4N_{d}$ & $32 \times 32 \times 8N_{d}$ \\\cline{2-4}
&ROI Align ($5$) & $32 \times 32 \times 8N_{d}$ & $N_{\textrm{large}} \times 5 \times 5 \times 8N_{d}$ \\\cline{2-4}
&ROI Encoder ($5$) & $N_{\textrm{large}} \times 5 \times 5 \times 8N_{d}$ & $h~(N_{\textrm{large}} \times 4 \times 4 \times 4N_{d})$ \\\cline{2-4}
&Repeat ($4 \times 4$) &$c^{\textrm{obj}}~(N_{\textrm{large}} \times N_{g})$ & $N_{\textrm{large}} \times 4 \times 4 \times N_{g}$ \\\cline{2-4}
&Repeat ($4 \times 4$) &$e^{\textrm{g}}~(N_{\textrm{large}} \times N_{l})$ & $N_{\textrm{large}} \times 4 \times 4 \times N_{l}$ \\\cline{2-4}
&Concat - Conv $3 \times 3~(4N_{d})$ & $h, N_{\textrm{large}} \times 4 \times 4 \times N_{g}, N_{\textrm{large}} \times 4 \times 4 \times N_{l}$  & $hc$ ($N_{\textrm{large}} \times 4 \times 4 \times 4N_{d}$)\\\cline{2-4}
&Outlogits (unconditional loss) & $h$  & $N_{\textrm{large}}$\\\cline{2-4}
&Outlogits (conditional loss) & $hc$  & $N_{\textrm{large}}$\\\hline

\end{tabular}
\label{tab:objd_struct}
\end{table*}

\begin{table*}[h]
\footnotesize
\centering
\caption{The structure for patch-wise spectral normalized projection discriminators of Obj-GAN. $\overline{e}$ is output by $F^{\textrm{ca}}$}
\begin{tabular}[t]{|c|p{5cm}|c|c|}\hline
{Stage} &{Name} &{\makebox[4cm]{Input Tensors}} &{\makebox[4cm]{Output Tensors}}\\ \hline\hline
\multirow{7}{*}{$D_{0}$} &Downsampling w/ SN ($N_{d}$) & $x_{0}~(64 \times 64 \times 3)$ & $32 \times 32 \times N_{d}$ \\\cline{2-4}
&Downsampling w/ SN ($2N_{d}$) & $32 \times 32 \times N_{d}$ & $16 \times 16 \times 2N_{d}$ \\\cline{2-4}
&Downsampling w/ SN ($4N_{d}$) & $16 \times 16 \times 2N_{d}$ & $8 \times 8 \times 4N_{d}$ \\\cline{2-4}
&Downsampling w/ SN ($8N_{d}$) & $8 \times 8 \times 4N_{d}$ & $4 \times 4 \times 8N_{d}$ \\\cline{2-4}
&Conv $4 \times 4$ w/ SN & $4 \times 4 \times 8N_{d}$  &$h_{0}~(8N_{d})$\\\cline{2-4}
&FC w/ SN ($8N_{d}$) &$\overline{e}~(N_{e})$ & $c_{0}~(8N_{d})$\\\cline{2-4}
&Fmap Mul - Avg Pool ($0$) &$h_{0}, c_{0}$ &$hc_{0}~(1)$\\\cline{2-4}
&Conv $1 \times 1$ w/ SN (unconditional loss) &$h_{0}$ &$o^{\textrm{uncond}}_{0}~(1)$\\\cline{2-4}
&Fmap Sum (conditional loss) &$o^{\textrm{uncond}}_{0},hc_{0}$ &$o^{\textrm{cond}}_{0}~(1)$\\\hline

\multirow{7}{*}{$D_{1}$} &Downsampling w/ SN ($N_{d}$) & $x_{1}~(128 \times 128 \times 3)$ & $64 \times 64 \times N_{d}$ \\\cline{2-4}
&Downsampling w/ SN ($2N_{d}$) & $64 \times 64 \times N_{d}$ & $32 \times 32 \times 2N_{d}$ \\\cline{2-4}
&Downsampling w/ SN ($4N_{d}$) & $32 \times 32 \times 2N_{d}$ & $16 \times 16 \times 4N_{d}$ \\\cline{2-4}
&Downsampling w/ SN ($8N_{d}$) & $16 \times 16 \times 4N_{d}$ & $8 \times 8 \times 8N_{d}$ \\\cline{2-4}
&Conv $4 \times 4$ w/ SN & $8 \times 8 \times 8N_{d}$  &$h_{1}~(3 \times 3 \times 8N_{d})$\\\cline{2-4}
&FC w/ SN ($8N_{d}$) &$\overline{e}~(N_{e})$ & $8N_{d}$\\\cline{2-4}
&Repeat ($3 \times 3$) &$8N_{d}$ &$c_{1}~(3 \times 3 \times 8N_{d})$ \\\cline{2-4}
&Fmap Mul - Avg Pool ($2$) &$h_{1}, c_{1}$ &$hc_{1}~(3 \times 3)$\\\cline{2-4}
&Conv $1 \times 1$ w/ SN (unconditional loss) &$h_{1}$ &$o^{\textrm{uncond}}_{1}~(3 \times 3)$\\\cline{2-4}
&Fmap Sum (conditional loss) &$o^{\textrm{uncond}}_{1},hc_{1}$ &$o^{\textrm{cond}}_{1}~(3 \times 3)$\\\hline

\multirow{7}{*}{$D_{2}$} &Downsampling w/ SN ($N_{d}$) & $x_{2}~(256 \times 256 \times 3)$ & $128 \times 128 \times N_{d}$ \\\cline{2-4}
&Downsampling w/ SN ($2N_{d}$) & $128 \times 128 \times N_{d}$ & $64 \times 64 \times 2N_{d}$ \\\cline{2-4}
&Downsampling w/ SN ($4N_{d}$) & $64 \times 64 \times 2N_{d}$ & $32 \times 32 \times 4N_{d}$ \\\cline{2-4}
&Downsampling w/ SN ($8N_{d}$) & $32 \times 32 \times 4N_{d}$ & $16 \times 16 \times 8N_{d}$ \\\cline{2-4}
&Conv $4 \times 4$ w/ SN & $16 \times 16 \times 8N_{d}$  &$h_{2}~(7 \times 7 \times 8N_{d})$\\\cline{2-4}
&FC w/ SN ($8N_{d}$) &$\overline{e}~(N_{e})$ & $8N_{d}$\\\cline{2-4}
&Repeat ($7 \times 7$) &$8N_{d}$ &$c_{2}~(7 \times 7 \times 8N_{d})$ \\\cline{2-4}
&Fmap Mul - Avg Pool ($2$) &$h_{2}, c_{2}$ &$hc_{2}~(7 \times 7)$\\\cline{2-4}
&Conv $1 \times 1$ w/ SN (unconditional loss) &$h_{2}$ &$o^{\textrm{uncond}}_{2}~(7 \times 7)$\\\cline{2-4}
&Fmap Sum (conditional loss) &$o^{\textrm{uncond}}_{2},hc_{2}$ &$o^{\textrm{cond}}_{2}~(7 \times 7)$\\\hline

\end{tabular}
\label{tab:pjpatd_struct}
\end{table*}

\begin{table*}[h]
\footnotesize
\centering
\vspace{-0.3cm}
\caption{The structure for shape spectral normalized projection discriminators of Obj-GAN.}
\vspace{-0.3cm}
\begin{tabular}[t]{|c|p{5cm}|c|c|}\hline
{Stage} &{Name} &{\makebox[4cm]{Input Tensors}} &{\makebox[4cm]{Output Tensors}}\\ \hline\hline
\multirow{7}{*}{$D_{0}$} &Shape Encoder w/ SN ($\frac{1}{8}N_{d}$) & $M^{0}$ ($64 \times 64 \times N_{c}$) &$64 \times 64\times \frac{1}{8}N_{d}$ \\\cline{2-4}
&Concat & $x_{0}~(64 \times 64 \times 3), 64 \times 64\times \frac{1}{8}N_{d}$ & $64 \times 64 \times (3+\frac{1}{8}N_{d})$ \\\cline{2-4}
&Downsampling w/ SN ($N_{d}$) & $64 \times 64 \times (3+\frac{1}{8}N_{d})$ & $32 \times 32 \times N_{d}$ \\\cline{2-4}
&Downsampling w/ SN ($2N_{d}$) & $32 \times 32 \times N_{d}$ & $16 \times 16 \times 2N_{d}$ \\\cline{2-4}
&Downsampling w/ SN ($4N_{d}$) & $16 \times 16 \times 2N_{d}$ & $8 \times 8 \times 4N_{d}$ \\\cline{2-4}
&Downsampling w/ SN ($8N_{d}$) & $8 \times 8 \times 4N_{d}$ & $4 \times 4 \times 8N_{d}$ \\\cline{2-4}
&Conv $4 \times 4$ w/ SN & $4 \times 4 \times 8N_{d}$  &$h_{0}~(8N_{d})$\\\cline{2-4}
&Conv $1 \times 1$ w/ SN (unconditional loss) &$h_{0}$ &$1$\\\hline

\multirow{7}{*}{$D_{1}$} &Shape Encoder w/ SN ($\frac{1}{8}N_{d}$) & $M^{1}$ ($128 \times 128 \times N_{c}$) &$128 \times 128 \times \frac{1}{8}N_{d}$ \\\cline{2-4}
&Concat & $x_{1}~(128 \times 128 \times 3), 128 \times 128 \times \frac{1}{8}N_{d}$ & $128 \times 128 \times (3+\frac{1}{8}N_{d})$ \\\cline{2-4}
&Downsampling w/ SN ($N_{d}$) & $128 \times 128 \times (3+\frac{1}{8}N_{d})$ & $64 \times 64 \times N_{d}$ \\\cline{2-4}
&Downsampling w/ SN ($2N_{d}$) & $64 \times 64 \times N_{d}$ & $32 \times 32 \times 2N_{d}$ \\\cline{2-4}
&Downsampling w/ SN ($4N_{d}$) & $32 \times 32 \times 2N_{d}$ & $16 \times 16 \times 4N_{d}$ \\\cline{2-4}
&Downsampling w/ SN ($8N_{d}$) & $16 \times 16 \times 4N_{d}$ & $8 \times 8 \times 8N_{d}$ \\\cline{2-4}
&Conv $4 \times 4$ w/ SN & $8 \times 8 \times 8N_{d}$  &$h_{1}~(3 \times 3 \times 8N_{d})$\\\cline{2-4}
&Conv $1 \times 1$ w/ SN (unconditional loss) &$h_{1}$ &$3 \times 3$\\\hline

\multirow{7}{*}{$D_{2}$} &Shape Encoder w/ SN ($\frac{1}{8}N_{d}$) & $M^{2}$ ($256 \times 256 \times N_{c}$) &$256 \times 256 \times \frac{1}{8}N_{d}$ \\\cline{2-4}
&Concat & $x_{2}~(256 \times 256 \times 3), 256 \times 256 \times \frac{1}{8}N_{d}$ & $256 \times 256 \times (3+\frac{1}{8}N_{d})$ \\\cline{2-4}
&Downsampling w/ SN ($N_{d}$) & $256 \times 256 \times (3+\frac{1}{8}N_{d})$ & $128 \times 128 \times N_{d}$ \\\cline{2-4}
&Downsampling w/ SN ($2N_{d}$) & $128 \times 128 \times N_{d}$ & $64 \times 64 \times 2N_{d}$ \\\cline{2-4}
&Downsampling w/ SN ($4N_{d}$) & $64 \times 64 \times 2N_{d}$ & $32 \times 32 \times 4N_{d}$ \\\cline{2-4}
&Downsampling w/ SN ($8N_{d}$) & $32 \times 32 \times 4N_{d}$ & $16 \times 16 \times 8N_{d}$ \\\cline{2-4}
&Conv $4 \times 4$ w/ SN & $16 \times 16 \times 8N_{d}$  &$h_{2}~(7 \times 7 \times 8N_{d})$\\\cline{2-4}
&Conv $1 \times 1$ w/ SN (unconditional loss) &$h_{2}$ &$7 \times 7$\\\hline

\end{tabular}
\label{tab:pjshpd_struct}
\end{table*}

\begin{table*}[h]
\footnotesize
\centering
\vspace{-0.3cm}
\caption{The structure for object-wise spectral normalized projection discriminators of Obj-GAN. $c^{\textrm{obj}}$ represents the intermediate context vectors of $F^{\textrm{obj-attn}}$, and $e^{\textrm{g}}$ represents the embedding vectors the class labels.}
\vspace{-0.3cm}
\begin{tabular}[t]{|c|p{5cm}|c|c|}\hline
{Stage} &{Name} &{\makebox[4cm]{Input Tensors}} &{\makebox[4cm]{Output Tensors}}\\ \hline\hline
\multirow{12}{*}{small} &Interpolating ($2$) & $M^{2}$ ($256 \times 256 \times N_{c}$) &$512 \times 512 \times N_{c}$ \\\cline{2-4}
&Interpolating ($2$) & $x^{2}$ ($256 \times 256 \times 3$) &$512 \times 512 \times 3$ \\\cline{2-4}
&Shape Encoder w/ SN ($\frac{1}{8}N_{d}$) & $512 \times 512 \times N_{c}$ &$512 \times 512 \times \frac{1}{8}N_{d}$ \\\cline{2-4}
&Concat & $512 \times 512 \times 3, 512 \times 512 \times \frac{1}{8}N_{d}$ & $512 \times 512 \times (3+\frac{1}{8}N_{d})$ \\\cline{2-4}
&Downsampling w/ SN ($N_{d}$) & $512 \times 512 \times (3+\frac{1}{8}N_{d})$ & $256 \times 256 \times N_{d}$ \\\cline{2-4}
&Downsampling w/ SN ($2N_{d}$) & $256 \times 256 \times N_{d}$ & $128 \times 128 \times 2N_{d}$ \\\cline{2-4}
&Downsampling w/ SN ($4N_{d}$) & $128 \times 128 \times 2N_{d}$ & $64 \times 64 \times 4N_{d}$ \\\cline{2-4}
&ROI Align ($5$) & $64 \times 64 \times 4N_{d}$ & $N_{\textrm{small}} \times 5 \times 5 \times 4N_{d}$ \\\cline{2-4}
&ROI Encoder w/SN ($5$) & $N_{\textrm{small}} \times 5 \times 5 \times 4N_{d}$ & $N_{\textrm{small}} \times 4 \times 4 \times 4N_{d}$ \\\cline{2-4}
&Conv $4 \times 4$ w/ SN & $N_{\textrm{small}} \times 4 \times 4 \times 4N_{d}$  &$h~(N_{\textrm{small}} \times 4N_{d})$\\\cline{2-4}
&Concat &$c^{\textrm{obj}}~(N_{\textrm{small}} \times N_{g}), e^{\textrm{g}}~(N_{\textrm{small}} \times N_{l})$ &$N_{\textrm{small}} \times (N_{g}+N_{l})$\\\cline{2-4}
&FC w/ SN ($4N_{d}$) & $N_{\textrm{small}} \times (N_{g}+N_{l})$ & $c~(N_{\textrm{small}} \times 4N_{d})$\\\cline{2-4}
&Fmap Mul - Avg Pool ($1$) &$h, c$ &$hc~(N_{\textrm{small}})$\\\cline{2-4}
&Conv $1 \times 1$ w/ SN (unconditional loss) &$h$ &$o^{\textrm{uncond}}~(N_{\textrm{small}})$\\\cline{2-4}
&Fmap Sum (conditional loss) &$o^{\textrm{uncond}},hc$ &$o^{\textrm{cond}}~(N_{\textrm{small}})$\\\hline\hline

\multirow{13}{*}{large} &Interpolating ($2$) & $M^{2}$ ($256 \times 256 \times N_{c}$) &$512 \times 512 \times N_{c}$ \\\cline{2-4}
&Interpolating ($2$) & $x^{2}$ ($256 \times 256 \times 3$) &$512 \times 512 \times 3$ \\\cline{2-4}
&Shape Encoder w/ SN ($\frac{1}{8}N_{d}$) & $512 \times 512 \times N_{c}$ &$512 \times 512 \times \frac{1}{8}N_{d}$ \\\cline{2-4}
&Concat & $512 \times 512 \times 3, 512 \times 512 \times \frac{1}{8}N_{d}$ & $512 \times 512 \times (3+\frac{1}{8}N_{d})$ \\\cline{2-4}
&Downsampling w/ SN ($N_{d}$) & $512 \times 512 \times (3+\frac{1}{8}N_{d})$ & $256 \times 256 \times N_{d}$ \\\cline{2-4}
&Downsampling w/ SN ($2N_{d}$) & $256 \times 256 \times N_{d}$ & $128 \times 128 \times 2N_{d}$ \\\cline{2-4}
&Downsampling w/ SN ($4N_{d}$) & $128 \times 128 \times 2N_{d}$ & $64 \times 64 \times 4N_{d}$ \\\cline{2-4}
&Downsampling w/ SN ($8N_{d}$) & $64 \times 64 \times 4N_{d}$ & $32 \times 32 \times 8N_{d}$ \\\cline{2-4}
&ROI Align ($5$) & $32 \times 32 \times 8N_{d}$ & $N_{\textrm{large}} \times 5 \times 5 \times 8N_{d}$ \\\cline{2-4}
&ROI Encoder w/ SN ($5$) & $N_{\textrm{large}} \times 5 \times 5 \times 8N_{d}$ & $N_{\textrm{large}} \times 4 \times 4 \times 4N_{d}$ \\\cline{2-4}
&Conv $4 \times 4$ w/ SN & $N_{\textrm{large}} \times 4 \times 4 \times 4N_{d}$  &$h~(N_{\textrm{large}} \times 4N_{d})$\\\cline{2-4}
&Concat &$c^{\textrm{obj}}~(N_{\textrm{large}} \times N_{g}), e^{\textrm{g}}~(N_{\textrm{large}} \times N_{l})$ &$N_{\textrm{large}} \times (N_{g}+N_{l})$\\\cline{2-4}
&FC w/ SN ($4N_{d}$) & $N_{\textrm{large}} \times (N_{g}+N_{l})$ & $c~(N_{\textrm{large}} \times 4N_{d})$\\\cline{2-4}
&Fmap Mul - Avg Pool ($1$) &$h, c$ &$hc~(N_{\textrm{large}})$\\\cline{2-4}
&Conv $1 \times 1$ w/ SN (unconditional loss) &$h$ &$o^{\textrm{uncond}}~(N_{\textrm{large}})$\\\cline{2-4}
&Fmap Sum (conditional loss) &$o^{\textrm{uncond}},hc$ &$o^{\textrm{cond}}~(N_{\textrm{large}})$\\\hline

\end{tabular}
\label{tab:pjobjd_struct}
\end{table*}

\end{document}